\documentclass[10pt,twocolumn,twoside]{IEEEtran}
\usepackage{indentfirst}
\usepackage{mathrsfs}
\usepackage{graphicx}
\usepackage{epstopdf}
\usepackage{amsmath, amssymb, amsthm, bm}
\usepackage{leftidx}
\usepackage{extarrows}
\usepackage{stfloats}
\usepackage{picinpar}
\usepackage{enumerate}

\usepackage{algorithm}
\usepackage{algorithmic}

\usepackage{epsfig}
\usepackage{latexsym}
\usepackage{amsfonts}
\usepackage{enumerate}
\usepackage{graphics}
\usepackage{float}
\usepackage{pict2e}
\usepackage{tikz}
\usepackage{lineno}
\usepackage[pdftex,colorlinks,linkcolor=blue,anchorcolor=blue,citecolor=blue]{hyperref}
\usepackage{moreverb}
\usepackage{color}
\usepackage[nosort]{cite}
\usepackage{subfigure}
\usepackage{multirow}
\usetikzlibrary{arrows,automata}
\usepackage{colortbl}
\usepackage{wrapfig}
\usepackage{threeparttable}

\usepackage{cuted}
\usepackage{units}
\usepackage{bbm}
\usepackage{bbding}
\usetikzlibrary{arrows,automata}

\definecolor{myGray}{rgb}{.95,.95,.95}

\newfont{\bb}{msbm10}

\newcommand{\tr}{^{\sf T}}
\newcommand{\M}[1]{{\bf{#1}}}

\newcommand{\m}[1]{{\mathrm{#1}}}

\allowdisplaybreaks[4]

\newtheorem{thm}{Theorem}
\newtheorem{ass}{Assumption}
\newtheorem{cor}{Corollary}
\newtheorem{lem}{Lemma}

\ifCLASSINFOpdf

\else

\fi

\newcommand{\Exp}[1]{\mathbb{E}\!\left[ #1 \right]}

\begin{document}

\title{Achieving Linear Speedup with ProxSkip in Distributed Stochastic Optimization}

\author{Luyao Guo,
Sulaiman A. Alghunaim,
Kun Yuan,
Laurent Condat, \IEEEmembership{Senior Member, IEEE},\\
Jinde Cao, \IEEEmembership{Fellow, IEEE}
\thanks{This work was supported in part by the National Natural Science Foundation of China under Grant Nos. 62603479 and 62576098. (Corresponding author: Jinde Cao.)}

\thanks{Luyao Guo is with the School of Computer Science and Engineering, Suzhou University of Technology, Suzhou 215500, China
(e-mail: \href{mailto:lyguo@szut.edu.cn}{{lyguo@szut.edu.cn}}).}
\thanks{Sulaiman A. Alghunaim is with the Department of Electrical Engineering, Kuwait University, Kuwait 13060, Kuwait
(e-mail: \href{mailto:sulaiman.alghunaim@ku.edu.kw}{{sulaiman.alghunaim@ku.edu.kw}}).}
\thanks{Kun Yuan is with Center for Machine Learning Research, Peking University, Beijing 100871, China
(e-mail: \href{mailto:kunyuan@pku.edu.cn}{{kunyuan@pku.edu.cn}}).}
\thanks{Laurent Condat is with King Abdullah University of Science and Technology, Thuwal 23955-6900, Saudi Arabia.}
\thanks{Jinde Cao is with the School of Mathematics, Southeast University, Nanjing 210096, China, and also with Purple Mountain Laboratories, Nanjing 211111, China(e-mail: \href{mailto:jdcao@seu.edu.cn}{{jdcao@seu.edu.cn}}).}
}

\date{}
\maketitle

\begin{abstract}
The ProxSkip algorithm for distributed optimization is gaining increasing attention due to its effectiveness in reducing communication. However, existing analyses of ProxSkip are limited to the strongly convex setting and fail to achieve linear speedup with respect to the number of nodes. Key questions regarding its behavior in the non-convex setting and the achievability of linear speedup remain open. In this paper, we revisit decentralized ProxSkip and answer these questions affirmatively. We provide a unified convergence analysis for stochastic non-convex, convex, and strongly convex problems, revealing how gradient noise, local updates, network connectivity, and data heterogeneity jointly determine the convergence behavior. To the best of our knowledge, this is the first analysis showing that decentralized ProxSkip achieves linear speedup in the number of nodes under stochastic gradients. Moreover, our results demonstrate that local updates can effectively reduce communication frequency and improve communication efficiency.
\end{abstract}


\section{Introduction}
In this work, we consider the following distributed stochastic optimization problem over a group of agents $[n]:=\{1,2,\ldots,n\}$ connected over a network:
\begin{equation}\label{EQ:Problem1}
\begin{aligned}
&f^\star = \min_{\M{x}\in \mathbb{R}^d}\Big[f(\M{x}):=\frac{1}{n}\sum_{i=1}^{n}f_i(\M{x})\Big],\\
&\text{with }f_i(\M{x})=\mathbb{E}_{\xi_i\sim\mathcal{D}_i}[F_i(\M{x},\xi_i)],
\end{aligned}
\end{equation}
where $\{\mathcal{D}_i\}^n_{i=1}$ represent data distributions, which can be heterogeneous across $n$ nodes, $f_i:\mathbb{R}^d\rightarrow \mathbb{R}$ is a smooth local function accessed by node $i$. In this setup, a network of nodes (also referred to as agents, workers, or clients) collaboratively seeks to minimize the average of the nodes' objectives. Solving problem \eqref{EQ:Problem1} in a distributed manner has garnered considerable attention in recent years \cite{Nguyen2024,Lian2017,ZhouY2025,Koloskova2021,Alghunaim2022,Guo2023}. These methods do not rely on a central coordinator and require each node to communicate only with its neighbors. Nevertheless, distributed optimization algorithms may still face challenges arising from communication bottlenecks.

To reduce communication costs, many techniques have been proposed. These techniques include compressing models and gradients \cite{Toghani2022,Huang2025,Fatkhullin2024}, using acceleration schemes \cite{Huang2025MP,Sadiev2022,Kovalev2020,HuanLi2020,HuanLi2022,Hendrikx2021,Song2023}, and implementing local updates \cite{Lin2020ICLR,Woodworth2020ICML,Stich2019,JiaLiu2021}. By applying these strategies, it is possible to reduce the amount of information exchanged between different nodes during training, thereby improving the efficiency of distributed training setups.

In this work, we focus on using local updates to reduce communication frequency. In centralized settings, local-SGD/FedAvg \cite{Stich2019,Khaled2020,Wang2021} has emerged as one of the most widely adopted optimization methods that employ local updates. However, when dealing with heterogeneous data, Local-SGD/FedAvg encounters the challenge of ``client-drift." This phenomenon arises from the diversity of functions on each node, causing each client to converge towards the minima of its respective function $f_i$, which may be significantly distant from the global optimum $f^{\star}$. To tackle this issue, several algorithms have been proposed, including Scaffold \cite{Karimireddy2020}, Scaffold with momentum \cite{ZihengCheng2023}, FedLin \cite{Mitra2021}, FedPD \cite{FedPDZhang2021}, FedDyn \cite{FedDyn}, VRL-SGD \cite{VRL-SGD}, FedGATE \cite{FedGATE}, SCALLION/SCAFCOM \cite{SCALLION}. In distributed settings, local-DSGD has been introduced in \cite{Stich2020}. Similarly to local-SGD, it also encounters the issue of client-drift when dealing with heterogeneous data. To mitigate the drift in Local-DSGD, several algorithms have been proposed. Notably, gradient-tracking (GT) based approaches, such as local-GT \cite{Nguyen2022} and $K$-GT \cite{Liu2023}, have been developed. Additionally, algorithms based on ED/NIDS/D$^2$ \cite{Exact diffusion,NIDS,D2,Guo2022}, such as LED \cite{Alghunaim2023}, have been introduced. In contrast to these periodic local-update methods \cite{Karimireddy2020,Mitra2021,FedPDZhang2021,Stich2020,Nguyen2022,Liu2023,Alghunaim2023}, probabilistic local-update methods include ProxSkip \cite{ProxSkip} and its extensions, such as TAMUNA \cite{TAMUNA}, CompressedScaffnew \cite{CompressedScaffnew}, LoCoDL \cite{LoCoDL}, VR-ProxSkip \cite{VR-ProxSkip}, ODEProx \cite{T-ProxSkip}, and RandProx \cite{RandProx}.

ProxSkip is known to be robust to data heterogeneity and to exhibit linear convergence on distributed strongly convex problems in the absence of stochastic noise \cite{ProxSkip}. When the network is sufficiently well-connected, ProxSkip \cite{ProxSkip}  and its extensions \cite{TAMUNA,CompressedScaffnew,LoCoDL,VR-ProxSkip,T-ProxSkip,RandProx} are gaining increasing attention due to their proven benefits in accelerating communication complexity. When deploying ProxSkip within the context of distributed learning, it becomes imperative to comprehend its behavior on non-convex tasks and its susceptibility to stochastic noise. However, existing ProxSkip convergence analyses focus on convex settings, and the main limitation of existing analyses is their inability to prove linear speedup with respect to the number of nodes (Table \ref{Table-Comparison}). Notice that although \cite{T-ProxSkip} presents the ODEProx algorithm and gives a tighter analysis of ProxSkip in the strongly convex setting, this new analysis shares the same limitation as the original ProxSkip analysis, namely, the inability to achieve linear speedup. Achieving linear speedup is highly desirable for a distributed learning and optimization algorithm as it enables effective utilization of the massive parallelism inherent in large distributed stochastic optimization systems. Consequently, two fundamental open questions emerge:

\emph{(1) How does ProxSkip behave on non-convex tasks?}

\emph{(2) Can we establish a linear speedup bound for ProxSkip in the presence of stochastic noise?}

\begin{table*}[t]
\centering
\renewcommand\arraystretch{2}
\caption{Comparisons with existing convergence rates of ProxSkip for distributed optimization. $\sigma^2$ is the variance of the stochastic gradient. $\zeta_0:=\max\{1-\alpha\mu,1-p^2\}$, the definition of $\zeta_{\m{new}}$ and $\zeta$ can be found in \cite{T-ProxSkip} and Lemma \ref{THM-1-PER}, respectively. In the last row, fixed network and problem parameters and initialization factors are suppressed.}
\resizebox{\textwidth}{!}{
\begin{tabular}{ccccccc}
\hline
\multirow{2}{*}{\textbf{Reference}}&\multicolumn{3}{c}{ \textbf{convergence rate} }&\multirow{2}{*}{\textbf{decentralized}} &\multirow{2}{*}{\textbf{linear speedup}}  \\
\cline{2-4}
 & \multicolumn{1}{c}{\textbf{N-C}}& \multicolumn{1}{c}{\textbf{C}}&\multicolumn{1}{c}{\textbf{SC}}\\
\hline
\cite{TAMUNA,CompressedScaffnew,LoCoDL}&no results&no results&$\zeta_0^T$, $\sigma^2=0$&\textcolor[rgb]{0.7,0,0}{\XSolidBrush}&\textcolor[rgb]{0.7,0,0}{\XSolidBrush}\\
\cite{ProxSkip,VR-ProxSkip}&no results&no results&$\zeta_0^T+\mathcal{O}(\alpha\sigma^2)$&\textcolor[rgb]{0.7,0,0}{\XSolidBrush}&\textcolor[rgb]{0.7,0,0}{\XSolidBrush}\\
\cite{T-ProxSkip}&no results&no results&$\zeta_{\m{new}}^T+\mathcal{O}(\alpha\sigma^2)$&\textcolor[rgb]{0.7,0,0}{\XSolidBrush}&\textcolor[rgb]{0.7,0,0}{\XSolidBrush}\\
\cite{ProxSkip}&no results&no results&$\zeta^T$, $\sigma^2=0$&\textcolor[rgb]{0,0.6,0}{\Checkmark}&\textcolor[rgb]{0.7,0,0}{\XSolidBrush}\\
\cite{RandProx}&no results&  $\mathcal{O}\left(\frac{1}{\alpha T}\right)$&$\zeta^T$, $\sigma^2=0$&\textcolor[rgb]{0,0.6,0}{\Checkmark}&\textcolor[rgb]{0.7,0,0}{\XSolidBrush}\\
\rowcolor[gray]{0.9}
\hline
\textbf{this work}&$\mathcal{O}\left(\frac{1}{\alpha T}+\frac{\alpha\sigma^2}{n}+\alpha^2\sigma^2\right)$&$\mathcal{O}\left(\frac{1}{\alpha T}+\frac{\alpha\sigma^2}{n}+\alpha^2\sigma^2\right)$&
$\mathcal{O}\left((1-\alpha\mu/2)^T+\frac{\alpha\sigma^2}{n}+\alpha^2\sigma^2\right)$&\textcolor[rgb]{0,0.6,0}{\Checkmark} &\textcolor[rgb]{0,0.6,0}{\Checkmark} \\
\hline
\end{tabular}}
\label{Table-Comparison}
\end{table*}
In this paper, we revisit ProxSkip for distributed stochastic optimization and provide answers to both questions. Specifically, we develop a new analysis with a novel proof technique under non-convex (N-C), convex (C), and strongly convex (SC) settings. Through this analysis, we obtain several new results that are comparable to the bounds of state-of-the-art distributed algorithms while achieving linear speedup bounds. We highlight our contributions as follows:
\begin{itemize}
 \item We establish the non-asymptotic convergence rate under stochastic non-convex, convex, and strongly convex settings of ProxSkip for problem \eqref{EQ:Problem1}. In particular, we prove that ProxSkip at iteration $T$ converges with rate
\begin{align*}
\text{(N-)C: }&\mathcal{O}\left(\frac{\sigma}{\sqrt{nT}}+\frac{\chi^{\nicefrac{1}{3}} \sigma^{\nicefrac{2}{3}}}{p^{\nicefrac{1}{3}}\rho^{\nicefrac{1}{3}} T^{\nicefrac{2}{3}}}
+\frac{\chi}{p\rho T}\right),\\
\text{SC: }&\tilde{\mathcal{O}}\left(\frac{\sigma^2}{n T}+\frac{\sigma^2\chi}{p\rho T^2}+\mathrm{exp}\left[-\nicefrac{cp\rho T}{\chi}\right]\right),
\end{align*}
where $\sigma^2$ denotes the variance of the stochastic gradient, $\rho:=1-\lambda_2$ is a topology-dependent quantity that approaches $0$ for a large and sparse network; $\chi\geq4$ and $c>0$ depends only on $L$ and $\mu$. To the best of our knowledge, this is the first work to establish convergence rates for a distributed method with probabilistic local updates in the non-convex setting.

\item We prove that, after enough transient time, the expected communication complexity of ProxSkip is $\mathcal{O}(\nicefrac{p\sigma^2}{n\epsilon^2})$ ( $\tilde{\mathcal{O}}(\nicefrac{p\sigma^2}{n\epsilon})$ for SC), where $\epsilon$ denotes the desired accuracy level, demonstrating that ProxSkip achieves linear speedup with respect to the number of nodes $n$. The proposed new proof technique overcomes the analytical limitations of \cite{ProxSkip,TAMUNA,CompressedScaffnew,LoCoDL,VR-ProxSkip,T-ProxSkip,RandProx}. To the best of our knowledge, we prove for the first time that ProxSkip can achieve linear speedup.

\item We elucidate the effects of noise, local steps, network topology, and data heterogeneity on the convergence of ProxSkip in stochastic non-convex, convex, and strongly convex settings. We demonstrate the robustness of ProxSkip against data heterogeneity while enhancing communication efficiency by local updates. Furthermore, we show that the convergence rates exhibited by ProxSkip in stochastic settings are comparable with those of existing state-of-the-art distributed algorithms incorporating local updates \cite{Stich2020,HuangYan2023,Nguyen2022,Liu2023,Alghunaim2023} (Table \ref{Table-Comparison-2}).
\end{itemize}

\section{Setup}\label{SEC-II}
Let $\mathbf{x}_i^t \in \mathbb{R}^d$ represent the local state of node $i$ at the $t$-th iteration. For the sake of convenience in notation, we use bold capital letters to denote stacked variables. For instance,
\begin{align*}
\mathbf{X}^t :=&\ [\mathbf{x}_{1}^t,\mathbf{x}_{2}^t,\ldots,\mathbf{x}_{n}^t]\tr \in \mathbb{R}^{n \times d} ,\\
\mathbf{G}^t := & \ [\mathbf{g}_1^t,\mathbf{g}_2^t,\ldots,\mathbf{g}_n^t]\tr \in \mathbb{R}^{n \times d},\\
\nabla F(\mathbf{X}^t) :=& \ [\nabla f_1(\mathbf{x}^t_1),\nabla f_2(\mathbf{x}^t_2),\ldots,\nabla f_n(\mathbf{x}^t_n)]\tr \in \mathbb{R}^{n \times d}.
\end{align*}

\subsection{Network Graph}
In this work, we focus on distributed scenarios with an undirected and connected network, where a network of $n$ nodes is interconnected by a graph with a set of edges $\mathcal{E}$, where node $i$ is connected to node $j$ if $(i,j) \in \mathcal{E}$. To describe the algorithm, we introduce the global mixing matrix $\mathbf{W}=[W_{ij}]$, where $W_{ij}=W_{ji}=0$ if $(i,j) \notin \mathcal{E}$, and $W_{ij}>0$ otherwise. We impose the following standard assumption on $\M{W}$.
\begin{ass}\label{MixingMatrix}
The mixing matrix $\mathbf{W}\in[0,1]^{n\times n}$ is symmetric, doubly stochastic, and primitive. Let $\lambda_1=1$ denote the largest eigenvalue of the mixing matrix $\mathbf{W}$, and the remaining eigenvalues are denoted as $1>\lambda_2\geq\lambda_3\geq\cdots\geq\lambda_n>-1$.
\end{ass}
We introduce two quantities as follows:
$$\mathbf{W}_a = \mathbf{I} - \nicefrac{1}{2\chi}(\mathbf{I} - \mathbf{W}),\ \mathbf{W}_b = (\mathbf{I} - \mathbf{W})^{1/2},$$
where $\chi \geq 1$. Under Assumption \ref{MixingMatrix}, the matrix $\mathbf{W}_a$ is positive semi-definite and doubly stochastic. Furthermore, we have $\mathbf{I} - \mathbf{W}_a = \nicefrac{1}{2\chi}\mathbf{W}_b^2$, and $\mathbf{W}_a$ is well-conditioned when $\chi$ is large.
\subsection{Algorithm Description}
Recalling \cite{ProxSkip}, the ProxSkip algorithm for the distributed stochastic optimization problem \eqref{EQ:Problem1} can be written as
\begin{equation}\label{Update:RandCom}
\begin{aligned}
\hat{\mathbf{Z}}^t &= \mathbf{X}^t - \alpha \mathbf{G}^t - \mathbf{Y}^t,\\
\mathbf{X}^{t+1} &= (1 - \theta_t)\hat{\mathbf{Z}}^t + \theta_t \mathbf{W}_a\hat{\mathbf{Z}}^t,\\
\mathbf{Y}^{t+1} &= \mathbf{Y}^t + p(\hat{\mathbf{Z}}^t - \mathbf{X}^{t+1}).
\end{aligned}
\end{equation}
Here, $\alpha>0$ is the stepsize, $\mathbf{g}_i^t$ is the stochastic gradient at node $i$,
and $\mathbf{Y}^t$ is the control variate. Communication occurs when
$\theta_t=1$, where $\{\theta_t\}_{t\geq0}$ is a common i.i.d.\
Bernoulli sequence with $\mathbb{P}(\theta_t=1)=p$, independent of the
initial variables and stochastic sampling. Although the sequence may
be generated in advance, $\theta_t$ is revealed to the analysis
filtration only at iteration $t$. The node-wise implementation is
given in Algorithm~\ref{RandCom}.

\begin{algorithm}[t!]
	\caption{ProxSkip for distributed stochastic optimization}
	\label{RandCom}
	\begin{algorithmic}[1]
		\STATE Input $\alpha > 0$, $0<p\leq1$, $\chi\geq1$, initial $\M{x}^0_{i} =\M{x}^0\in\mathbb{R}^d$ and $\M{y}^0_i = 0,~i=1,\dots,n$, weights for averaging $\M{W}_a=\M{I}-\nicefrac{1}{2\chi}(\M{I-W}):=(\widetilde{W}_{ij})^n_{i,j=1}$.
        \STATE Generate a common i.i.d.\ Bernoulli sequence
$\{\theta_t\}_{t=0}^{T-1}$ with $\mathbb{P}(\theta_t=1)=p$.
		\FOR{$t=0,1,\dotsc,T-1$ every node}
        \STATE Sample $\xi_i^{t}$, compute gradient $\M{g}_i^t=\nabla F_i(\M{x}^t_i,\xi^{t}_i)$
		\STATE  $\hat{\M{z}}^t_i=\M{x}^t_i-\alpha\M{g}^t_i- \M{y}^t_i$
		\IF{$\theta_t=1$}
        \STATE $\M{x}^{t+1}_i=\sum_{j=1}^{n}\widetilde{W}_{ij}\hat{\M{z}}^t_j$,\ $\M{y}^{t+1}_i=\M{y}^{t}_i+p(\hat{\M{z}}^t_i-\M{x}^{t+1}_i)$
		\ELSE
		\STATE $\M{y}^{t+1}_i =  \M{y}^{t}_i,~\M{x}^{t+1}_i=\hat{\M{z}}^{t}_i$
		\ENDIF
		\ENDFOR
	\end{algorithmic}
\end{algorithm}

\subsection{Assumptions}
We further use the following standard assumptions:
\begin{ass}\label{ASS1}
A solution exists to problem \eqref{EQ:Problem1}, and $f^*>-\infty$. Moreover, $f_i$ is $L$-smooth, i.e.,
$\|\nabla f_i(\M{x})-\nabla f_i(\M{y})\|\leq L \|\M{x}-\M{y}\|, \forall \M{x},\M{y}\in \mathbb{R}^d$.
\end{ass}

\begin{ass}\label{Strongly-convex-ASS}
Each function $f_i$ is $\mu$-strongly convex for constant $\mu\geq0$, i.e.,
$
f_i(\M{x})-f_i(\M{y})+\frac{\mu}{2}\|\M{x}-\M{y}\|^2\leq\langle\nabla f_i(\M{x}),\M{x}-\M{y}\rangle,   \forall \M{x},\M{y}\in \mathbb{R}^d$.
\end{ass}

Let $\xi^t=(\xi_1^t,\ldots,\xi_n^t)$. For any collection of random
variables, $\sigma(\cdot)$ denotes the sigma-algebra generated by that
collection. We define the pre-sampling analysis history by $\mathcal{F}^t
:=
\sigma\!\left(
\M{X}^0,\M{Y}^0,
\{\xi^k,\theta_k\}_{k=0}^{t-1}
\right)$.
Thus, $(\M{X}^t,\M{Y}^t)$ is
$\mathcal{F}^t$-measurable.

\begin{ass}\label{StochasticGradient1}
Conditioned on $\mathcal{F}^t$, the samples
$\{\xi_i^t\}_{i=1}^n$ are mutually independent. Moreover, for every
$i=1,\ldots,n$ and $t\geq0$, the stochastic gradient
$\mathbf{g}_i^t=\nabla F_i(\mathbf{x}_i^t,\xi_i^t)$ satisfies
$
\mathbb{E}\!\left[
\mathbf{g}_i^t\mid\mathcal{F}^t
\right]
=
\nabla f_i(\mathbf{x}_i^t),
$
and there exists a constant $\sigma\geq0$ such that, almost surely,
$
\frac{1}{n}\sum_{i=1}^n
\mathbb{E}\!\left[
\left\|
\mathbf{g}_i^t-\nabla f_i(\mathbf{x}_i^t)
\right\|^2
\mid\mathcal{F}^t
\right]
\leq\sigma^2
$.
\end{ass}

\section{Main Convergence Results}\label{SEC-III}
We now present our novel convergence results for ProxSkip. In Section \ref{SEC-Pre}, we recall the existing results in \cite{ProxSkip}. In Section \ref{sec:mt},
the convergence rates and communication complexities are presented in Theorem \ref{MainResult-Convergence-rate-T}, Corollary \ref{MainResult-Convergence-rate-tighter-stepsize}, and Corollary \ref{MainResult-Convergence-rate-C}, respectively.

\subsection{Preliminary}\label{SEC-Pre}
We first recall existing convergence results for ProxSkip \cite{ProxSkip,RandProx}.
\begin{lem}\label{THM-1-PER}
Suppose that Assumptions \ref{MixingMatrix}, \ref{ASS1}, and \ref{StochasticGradient1} hold, and $f_i$ is $\mu$-strongly convex for some $0<\mu\leq L$. If $0<\alpha\leq\nicefrac{1}{L}$ and $\chi\geq1$, it holds that
\begin{align}\label{ProxSkip-Result}
\Exp{\big\|\bar{\M{x}}^{t+1}-\M{x}^{\star}\big\|^2}\leq  \zeta^{t+1}a_0+\frac{\alpha^2\sigma^2}{1-\zeta},
\end{align}
where $a_0$ is a constant that depends on the initialization and $\zeta=\max\left\{1-\alpha\mu,1-\frac{(1-\lambda_2)p^2}{2\chi} \right\}<1$.
\end{lem}
When $\sigma^2=0$, by setting $\alpha=\nicefrac{1}{L}$ and $\chi=1$, we can deduce from \eqref{ProxSkip-Result} that the communication complexity of ProxSkip to achieve $\epsilon$-accuracy, i.e., $\mathbb{E}[\|\bar{\M{x}}^{t}-\M{x}^{\star}\|^2]\leq\epsilon$, is given by
$
\mathcal{O}((p\kappa+\nicefrac{1}{p(1-\lambda_2)})\mathrm{log}\nicefrac{1}{\epsilon})$, where $\kappa=\nicefrac{L}{\mu}$. If the network is sufficiently well-connected, i.e., $\nicefrac{1}{(1-\lambda_2)\kappa}<1$, and we set $p=\sqrt{\nicefrac{1}{(1-\lambda_2)\kappa}}$, the communication complexity becomes $\mathcal{O}(\sqrt{\nicefrac{\kappa}{1-\lambda_2}}\ \mathrm{log}\nicefrac{1}{\epsilon})$, achieving the optimal communication complexity as proven by \cite{Scaman2017}. When $\sigma^2\neq0$, \eqref{ProxSkip-Result} bounds the mean-square error of the averaged iterate $\bar{\M{x}}^t$ by a geometric transient and $\alpha^2\sigma^2/(1-\zeta)$. For fixed network and communication parameters, this noise bound is $\mathcal O(\alpha\sigma^2)$ as $\alpha\to0$. Note that linear speedup is the property that the dominant term in the number of iterations $T$ needed to achieve $\epsilon$ accuracy is divided by $n$, e.g.,  $\frac{1}{n \epsilon^2}$ for nonconvex and $\frac{1}{n \epsilon}$ for SC, implying that increasing the number of nodes speeds up the convergence. The original  result in \cite{ProxSkip} cannot choose a stepsize to achieve this due to the term $O(\alpha \sigma^2)$ not divided by $n$. Therefore, further analysis is required to achieve the desired linear speedup.

\subsection{Main Theorem—Convergence Rate of ProxSkip}\label{sec:mt}
We are now ready to present the new convergence results for ProxSkip. The complete proofs are given in Sections \ref{AP:nonconvex}--\ref{AP:strongly-convex}, with the common error estimates established in Section \ref{AP:mainproof}.
\begin{thm}\label{MainResult-Convergence-rate-T}
Suppose that Assumptions \ref{MixingMatrix}, \ref{ASS1}, and \ref{StochasticGradient1} hold. Let $\bar{\M{x}}^t=\frac{1}{n}\sum_{i=1}^{n}\M{x}_i^t$ denote the iterates of Algorithm \ref{RandCom} and $\M{x}^{\star}$ solves \eqref{EQ:Problem1}. For $\chi\geq4$ and $0<\alpha\leq\frac{p(1-\lambda_2)}{320\chi L}$, we have the following convergence results.

\noindent\textbf{Non-convex:} Let $F_0=f(\bar{\M{x}}^0)-f^\star$ and $\varsigma^2_0=\frac{1}{n}\sum_{i=1}^{n}\|\nabla f_i(\bar{\M{x}}^0)-\nabla f(\bar{\M{x}}^0)\|^2$. It holds that
\begin{align}\label{NCVXMainResult1-T}
&\frac{1}{T}\sum_{t=0}^{T-1}\Exp{\big\|\nabla f(\bar{\M{x}}^t)\big\|^2}\nonumber\\
&\leq \mathcal{O}\bigg(\underbrace{\frac{F_0}{\alpha T}+\frac{\chi^2L^2\varsigma^2_0\alpha^2}{p^3(1-\lambda_2)^2T}}_{\m{deterministic\ part}}+\underbrace{\frac{\alpha\sigma^2L}{n}+\frac{\chi L^2\sigma^2\alpha^2}{p(1-\lambda_2)}}_{\m{stochastic\ part}}\bigg).
\end{align}

\noindent\textbf{Convex:} Let $R_0^2=\|\bar{\M{x}}^0-\M{x}^{\star}\|^2$. Under the additional Assumption \ref{Strongly-convex-ASS}, it holds that
\begin{align}\label{CVXMainResult2-T}
&\frac{1}{T}\sum_{t=0}^{T-1}\Exp{f(\bar{\M{x}}^t)-f^{\star}}\nonumber\\
&\leq \mathcal{O}\bigg(\underbrace{\frac{R_0^2}{\alpha T}+\frac{\chi^2L\varsigma^2_0\alpha^2}{p^3(1-\lambda_2)^2T}}_{\m{deterministic\ part}}+\underbrace{\frac{\alpha\sigma^2}{n}+\frac{\chi L\sigma^2\alpha^2}{p(1-\lambda_2)}}_{\m{stochastic\ part}}\bigg).
\end{align}

\noindent\textbf{Strongly convex:} Under the additional Assumption \ref{Strongly-convex-ASS} with $\mu>0$, it holds that
\begin{align}\label{SCVXMainResult-NEW-T}
\Exp{\big\|\bar{\M{x}}^{T}-\M{x}^{\star}\big\|^2}\leq&  \underbrace{\Big(1-\frac{\alpha\mu}{2}\Big)^T\Big(R_0^2+\frac{200\chi^2\alpha^3L\varsigma_0^2}{p^3(1-\lambda_2)^2}\Big)}_{\m{deterministic\ part}}\nonumber\\
&+\mathcal{O}\bigg(\underbrace{\frac{\alpha\sigma^2}{\mu n}+\frac{\chi L\sigma^2\alpha^2}{ p\mu(1-\lambda_2)}}_{\m{stochastic\ part}}\bigg).
\end{align}
\end{thm}
For the non-convex setting, Theorem \ref{MainResult-Convergence-rate-T} bounds the time-averaged squared gradient norm by a vanishing initialization term and a noise-dependent term. Without any additional assumptions, a stationary point is the best guarantee possible and is a satisfactory criterion to measure the performance of distributed methods with nonconvex objectives \cite{Stich2020}. For the convex case, it bounds the time-averaged objective gap. When $\sigma^2=0$, these two error criteria vanish at sublinear rates, while the mean-square distance to the optimum converges linearly in the strongly convex setting.

By choosing the stepsize carefully, Theorem \ref{MainResult-Convergence-rate-T} yields the following tighter rate.
\begin{table*}[!t]
\renewcommand\arraystretch{2}
\begin{center}
\caption{A comparison with existing methods employing local steps. $\rho=1-\lambda_2$, $K$ denotes the number of local steps.}
\begin{threeparttable}
\scalebox{1.25}{
\begin{tabular}{lllccc}
\hline
\multirow{2}{*}{\textbf{Method}}&\multicolumn{2}{c}{\# \textbf{communication rounds} } \\
\cline{2-3}
 & \multicolumn{1}{c}{\textbf{N-C/C}}&\multicolumn{1}{c}{\textbf{SC}}\\
\hline
local-DSGD \cite{Stich2020}&$\mathcal{O}\left(\frac{\sigma^2}{nK\epsilon^2}+\left(\frac{\sigma}{\sqrt{\rho K}}+\frac{\varsigma}{\rho}\right)\frac{1}{\epsilon^{\nicefrac{3}{2}}}+\frac{1}{\rho\epsilon}\right)$\tnote{a}~~~ &$\tilde{\mathcal{O}}\left(\frac{\sigma^2}{nK\epsilon}+\left(\frac{\sigma}{\sqrt{\rho K}}+\frac{\varsigma}{\rho}\right)\frac{1}{\sqrt{\epsilon}}+\frac{1}{\rho}\mathrm{log}\nicefrac{1}{\epsilon}\right)$\\
local-GT \cite{HuangYan2023,Nguyen2022}&
$\mathcal{O}\left(\frac{\sigma^2}{nK\epsilon^2}+\left(\frac{\sigma}{\rho^2\sqrt{K}}\right)\frac{1}{\epsilon^{\nicefrac{3}{2}}}+\frac{1}{\rho^2\epsilon}\right)\tnote{b}$&
$
\tilde{O}\left(\frac{\sigma^2}{nK\epsilon}+\frac{\sigma}{\sqrt{K}\rho}\frac{1}{\epsilon^{1/2}}+\frac{1}{\rho^2}\mathrm{log}\nicefrac{1}{\epsilon}\right)
$\\
$K$-GT \cite{Liu2023}&$\mathcal{O}\left(\frac{\sigma^2}{nK\epsilon^2}+\left(\frac{\sigma}{\rho^2\sqrt{K}}\right)\frac{1}{\epsilon^{\nicefrac{3}{2}}}+\frac{1}{\rho^2\epsilon}\right)\tnote{b}$
& no results\\
Periodical GT \cite{Liu2023}&$\mathcal{O}\left(\frac{\sigma^2}{nK\epsilon^2}+\left(\frac{\sigma}{\rho^2}\right)\frac{1}{\epsilon^{\nicefrac{3}{2}}}+\frac{1}{\rho^2\epsilon}\right)\tnote{b}$
& no results\\
LED \cite{Alghunaim2023}&$\mathcal{O}\left(\frac{\sigma^2}{nK\epsilon^2}+\left(\frac{\sigma}{\sqrt{\rho K}}\right)\frac{1}{\epsilon^{\nicefrac{3}{2}}}+\frac{1}{\rho\epsilon}\right)$&
$\tilde{\mathcal{O}}\left(\frac{\sigma^2}{nK\epsilon}+\left(\frac{\sigma}{\sqrt{\rho K}}\right)\frac{1}{\sqrt{\epsilon}}+\frac{1}{\rho}\mathrm{log}\nicefrac{1}{\epsilon}\right)$\\
\rowcolor[gray]{0.9}
\hline
\textbf{ProxSkip} &$\mathcal{O}\left(\frac{p\sigma^2}{n\epsilon^2}+\frac{\sqrt{p\chi}}{\sqrt{\rho}}\frac{\sigma }{\epsilon^{\nicefrac{3}{2}}}+\frac{\chi}{\rho\epsilon}\right)$&
$\tilde{\mathcal{O}}\left(\frac{p\sigma^2}{n\epsilon}+\frac{\sqrt{p\chi}}{\sqrt{\rho}}\frac{\sigma}{\sqrt{\epsilon}}+\frac{\chi}{\rho}\mathrm{log}\nicefrac{1}{\epsilon}\right)$\\
\hline
\end{tabular}}
 \begin{tablenotes}
  \footnotesize
   \item[a] $\varsigma^2$ is a function heterogeneity constant such that $\nicefrac{1}{n}\sum_{i=1}^{n}\|\nabla f_i(\M{x})-\nabla f(\M{x})\|^2\leq\varsigma^2$ for all $\M{x}$.
   \item[b] The result is for the non-convex setting, and no corresponding result is given for the convex setting.
 \end{tablenotes}
\end{threeparttable}
\label{Table-Comparison-2}
\end{center}
\end{table*}

\begin{cor}\label{MainResult-Convergence-rate-tighter-stepsize}
Under the same setting as in Theorem \ref{MainResult-Convergence-rate-T}, we have the following convergence results.\\
\noindent\textbf{Non-convex:} There exists a constant stepsize such that
\begin{align}\label{NCVXMainResult1-resp-new}
&\frac{1}{T}\sum_{t=0}^{T-1}\mathbb{E}[\|\nabla f(\bar{\M{x}}^t)\|^2]\nonumber\\
&\leq\mathcal{O}\left(\left(\frac{\sigma^2}{nT}\right)^{\frac{1}{2}}+\left(\frac{\chi \sigma^2}{p(1-\lambda_2)T^2}\right)^{\frac{1}{3}}
+\frac{\chi}{p(1-\lambda_2)T}\right).
\end{align}
\noindent\textbf{Convex:} Under the additional Assumption \ref{Strongly-convex-ASS} with $\mu\geq0$, there exists a constant stepsize such that
\begin{align}\label{CVXMainResult2-resp-new}
&\frac{1}{T}\sum_{t=0}^{T-1}\Exp{f(\bar{\M{x}}^t)-f^{\star}}\nonumber\\
&\leq\mathcal{O}\left(\left(\frac{\sigma^2}{nT}\right)^{\frac{1}{2}}+\left(\frac{\chi \sigma^2}{p(1-\lambda_2)T^2}\right)^{\frac{1}{3}}
+\frac{\chi}{p(1-\lambda_2)T}\right).
\end{align}
\noindent\textbf{Strongly Convex:} Under the additional Assumption \ref{Strongly-convex-ASS} with $\mu>0$, there exists a constant stepsize such that
\begin{align}\label{SCVXMainResult-NEW-resp-new}
&\mathbb{E}[\|\bar{\M{x}}^{T}-\M{x}^{\star}\|^2]\nonumber\\
&\leq\tilde{\mathcal{O}}\left(\frac{\sigma^2}{n T}+\frac{\sigma^2\chi}{p(1-\lambda_2) T^2}+\mathrm{exp}\Big[-\frac{cp(1-\lambda_2)T}{\chi}\Big]\right).
\end{align}
Here, $c=\mu/(640L)$. The $\mathcal{O}(\cdot)$ notation suppresses constants depending on $L$, $\mu$ when applicable, and the initial errors, but not on $n,p,\chi,\rho,T$, or $\sigma$; $\tilde{\mathcal{O}}(\cdot)$ additionally suppresses logarithmic factors.
\end{cor}

When $T$ is sufficiently large, the term $\frac{\sigma}{\sqrt{n T}}$ (and $\frac{\sigma^2}{n T}$ for SC) will dominate the rate. In this scenario, a sufficient iteration count for ProxSkip is $T=\mathcal{O}(\frac{\sigma^2}{n\epsilon^2})$ (and $T=\tilde{\mathcal{O}}(\frac{\sigma^2}{n\epsilon})$ for SC setting) iterations to reach a desired $\epsilon$-accurate solution, thus the convergence accuracy improves linearly with $n$.

It is crucial to note that while reducing $p$ leaves the first term of the convergence rate unchanged, it cannot be reduced indiscriminately. When $p$ is above a certain threshold, the first term (independent of $p$) dominates. Reducing $p$ in this regime has minimal impact on the rate, i.e., we can skip some communication almost for free, which explains why local updates can accelerate communication. However, if $p$ falls below this threshold, the second and third terms become dominant, and any further reduction in $p$ will directly slow down convergence.

\begin{cor}\label{MainResult-Convergence-rate-C}
Under the same setting as in Theorem \ref{MainResult-Convergence-rate-T} and Corollary \ref{MainResult-Convergence-rate-tighter-stepsize}, we have the following convergence results.

\noindent\textbf{Non-convex:} It holds that $\frac{1}{T}\sum_{t=0}^{T-1}\mathbb{E}[\|\nabla f(\bar{\M{x}}^t)\|^2]\leq\epsilon$ after
\begin{align}\label{NCVXMainResult1}
p\times \underbrace{\mathcal{O}\left(\frac{\sigma^2}{n\epsilon^2}+\frac{\sqrt{\chi}}{\sqrt{p}\sqrt{1-\lambda_2}}\frac{\sigma }{\epsilon^{\nicefrac{3}{2}}}+\frac{\chi}{p(1-\lambda_2)\epsilon}\right)}_{\m{iteration~rounds}}
\end{align}
expected communication rounds.

\noindent\textbf{Convex:} Under the additional Assumption \ref{Strongly-convex-ASS} with $\mu\geq0$, it holds that $\frac{1}{T}\sum_{t=0}^{T-1}\mathbb{E}[f(\bar{\M{x}}^t)-f^{\star}]\leq\epsilon$ after
\begin{align}\label{CVXMainResult2}
p\times
\underbrace{\mathcal{O}\left(\frac{\sigma^2}{n\epsilon^2}+\frac{\sqrt{\chi}}{\sqrt{p}\sqrt{1-\lambda_2}}\frac{\sigma }{\epsilon^{\nicefrac{3}{2}}}+\frac{\chi}{p(1-\lambda_2)\epsilon}\right)}_{\m{iteration~rounds}}
\end{align}
expected communication rounds.

\noindent\textbf{Strongly Convex:} Under the additional Assumption \ref{Strongly-convex-ASS} with $\mu>0$, it holds that $\mathbb{E}[\|\bar{\M{x}}^{T}-\M{x}^{\star}\|^2]\leq\epsilon$ after
\begin{align}\label{SCVXMainResult-NEW}
p\times
\underbrace{\tilde{\mathcal{O}}\left(\frac{\sigma^2}{n\epsilon}+\frac{\sqrt{\chi}}{\sqrt{p}\sqrt{1-\lambda_2}}\frac{\sigma}{\sqrt{\epsilon}}
+\frac{\chi\mathrm{log}\nicefrac{1}{\epsilon}}{p(1-\lambda_2)}\right)}_{\m{iteration~rounds}}
\end{align}
expected communication rounds. Here, the notation $\tilde{\mathcal{O}}(\cdot)$ ignores logarithmic factors.
\end{cor}

We provide Table \ref{Table-Comparison-2} to compare the convergence results of ProxSkip with existing state-of-the-art distributed optimization algorithms, such as local-DSGD \cite{Stich2020}, $K$-GT \cite{Liu2023}, and LED \cite{Alghunaim2023}, with local updates in terms of the number of communication rounds needed to achieve $\epsilon>0$.

\noindent\textbf{Achieving acceleration by $p$ and $n$.}  According to \eqref{NCVXMainResult1}, \eqref{CVXMainResult2}, and \eqref{SCVXMainResult-NEW}, when $\epsilon$ is sufficiently small, the convergence rate is dominated by noise and is unaffected by the graph parameter $1-\lambda_2$ for ProxSkip. After enough transient time, ProxSkip with $\mathcal{O}\big(\frac{p\sigma^2}{n\epsilon^2}\big)$ ($\tilde{\mathcal{O}}\big(\frac{p\sigma^2}{n \epsilon}\big)$ for SC) achieves linear speedup by the probability of communication $p$ and the number of nodes $n$. Note that reducing the communication probability $ p $ beyond a certain point is counterproductive. The communication bound retains the term $\mathcal{O}(\frac{\chi}{\rho\epsilon})$ ($\frac{\chi}{\rho}\mathrm{log}\nicefrac{1}{\epsilon}$ for SC), which is independent of $p$. Moreover, the stepsize condition $\alpha\leq\frac{p(1-\lambda_2)}{320\chi L}$ implies that an excessively small $p$ forces a small stepsize and can slow convergence.

\noindent\textbf{Removing dependence on data heterogeneity.} According to Table \ref{Table-Comparison-2}, the second term of the communication complexity of local-DSGD \cite{Stich2020}, a popular algorithm for distributed optimization, is as follows:
$\text{(N-)C: }\mathcal{O}\left(\left(\nicefrac{\sigma}{\sqrt{\rho K}}+\nicefrac{\varsigma}{\rho}\right)\epsilon^{-\nicefrac{3}{2}}\right)$,
$\text{SC: }\tilde{\mathcal{O}}\left(\left(\nicefrac{\sigma}{\sqrt{\rho K}}+\nicefrac{\varsigma}{\rho}\right)\epsilon^{-\nicefrac{1}{2}}\right)$.
Here, $\varsigma^2$ represents the function heterogeneity constant such that $\nicefrac{1}{n}\sum_{i=1}^{n}\|\nabla f_i(\mathbf{x})-\nabla f(\mathbf{x})\|^2\leq\varsigma^2$ for all $\mathbf{x}$. We note that ProxSkip lacks the additional term $\frac{\varsigma}{\rho}\frac{1}{\epsilon^{\nicefrac{3}{2}}}$ ($\frac{\varsigma}{\rho}\frac{1}{\sqrt{\epsilon}}$ for SC). Thus, ProxSkip avoids a heterogeneity-dependent steady-state error; data heterogeneity enters our bounds only through the initialization quantity $\varsigma_0^2$.

\noindent\textbf{Comparable with existing distributed algorithms incorporating local updates.} Choosing $\chi=4$, the non-convex and convex communication complexity of ProxSkip is
$
\mathcal{O}\left(\frac{p\sigma^2}{n\epsilon^2}+\frac{\sqrt{p}}{\sqrt{1-\lambda_2}}\frac{\sigma}{\epsilon^{\nicefrac{3}{2}}}+\frac{1}{(1-\lambda_2)\epsilon}\right)
$ for every $p\in(0,1]$. Thus, with $K=1/p$, ProxSkip matches the orders in Table \ref{Table-Comparison-2} for LED in both the non-convex/convex and strongly convex settings. Compared with GT based methods \cite{HuangYan2023,Nguyen2022,Liu2023}, the non-convex network dependence is improved.

\section{Experiments}\label{Sec-Experimental Results}
We empirically verify the theoretical results of ProxSkip for distributed stochastic optimization. We use connected networks with $n$ agents, with $\iota$ indicating network connectivity. The mixing matrix $\M{W}$ is generated with the Metropolis-Hastings rule. All stochastic results are averaged over 10 runs.

\subsection{Synthetic Dataset}
\begin{figure*}[!t]
\centering
\setlength{\abovecaptionskip}{0pt}
\includegraphics[width=\linewidth]{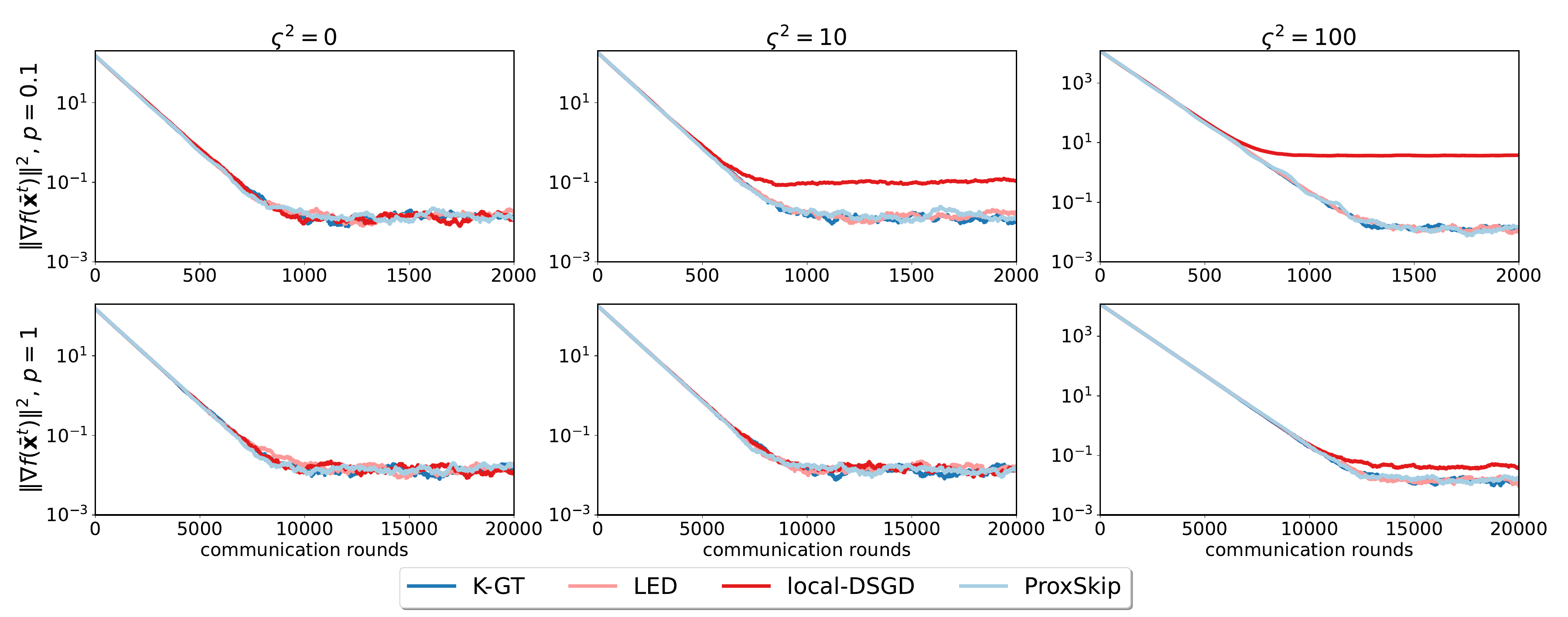}
\caption{Learning synthetic convex function over $10$ nodes with noise $\sigma^2=1$.}
\label{Sim:Convex-F1}
\end{figure*}

We construct the distributed least squares objective with $f_i(\M{x})=\nicefrac{1}{2}\|\M{A}_i\M{x}-\M{b}_i\|^2$ with fixed Hessian $\M{A}_i^2=\nicefrac{i^2}{n^2}\M{I}_d$, and sample each $\M{b}_i\thicksim\mathcal{N}(0,\nicefrac{\varsigma^2}{i^2}\M{I}_d)$ for each node $i\in[n]$, where $\varsigma^2$ can
control the deviation between local objectives \cite{Stich2020}. Stochastic noise is controlled by adding Gaussian noise with $\sigma^2=1$. We use a ring topology with $10$ nodes for this experiment. For all algorithms, we use the same stepsize $\alpha=0.001$. The results are shown in Fig. \ref{Sim:Convex-F1}. According to Fig. \ref{Sim:Convex-F1} the ``client-drift'' only happens for local-DSGD, where the larger $\varsigma^2\neq0$ gets, the poorer model quality D-SGD ends up with. From $\nicefrac{1}{p}=1$ to $\nicefrac{1}{p}=10$ in Fig. \ref{Sim:Convex-F1}, $K$-GT \cite{Liu2023}, LED \cite{Alghunaim2023}, and ProxSkip reach the same target after $10000$ rounds to only $1000$ achieving linear speedup in communication by local steps. However, more local steps makes local-DSGD suffer even more in model quality. This is consistent with the theoretical results from \cite{Liu2023} and \cite{Alghunaim2023}.

\subsection{Real-world Dataset ijcnn1}
The objective function is
$
f(\M{x})=\frac{1}{n}\sum_{i=1}^{n}\frac{1}{m_i}\sum_{j=1}^{m_{i}} \ln (1+e^{-(\mathcal{A}_{i j}\tr {\M{x}})\mathcal{B}_{i j}})+r(\M{x})
$.
Here, $r(\M{x})$ is the regularizer, node $i$ holds its own training data $\left(\mathcal{A}_{i j}, \mathcal{B}_{i j}\right) \in$ $\mathbb{R}^{d} \times\{-1,1\}, j=1, \cdots, m_{i}$, including sample vectors $\mathcal{A}_{i j}$ and corresponding classes $\mathcal{B}_{i j}$. We use the ijcnn1 dataset from the widely used LIBSVM library \cite{LibSVM}, which has $d=22$ attributes and $\sum_{i=1}^{n}m_i=49950$. Moreover, the training samples are randomly and evenly distributed over all the $n$ agents. We generate stochastic gradients as $\nabla F_i(\M{x})=\nabla f_i(\M{x})+\omega_i$, where $\omega_i\thicksim\mathcal{N}(0,10^{-3}\M{I}_d)$. Since $d=22$, the noise satisfies $\mathbb{E}\|\omega_i\|^2=22\times10^{-3}=0.022$, corresponding to $\sigma^2=0.022$ in Assumption~\ref{StochasticGradient1}.

\begin{figure*}[!t]
\centering
\setlength{\abovecaptionskip}{0pt}
\includegraphics[width=\linewidth]{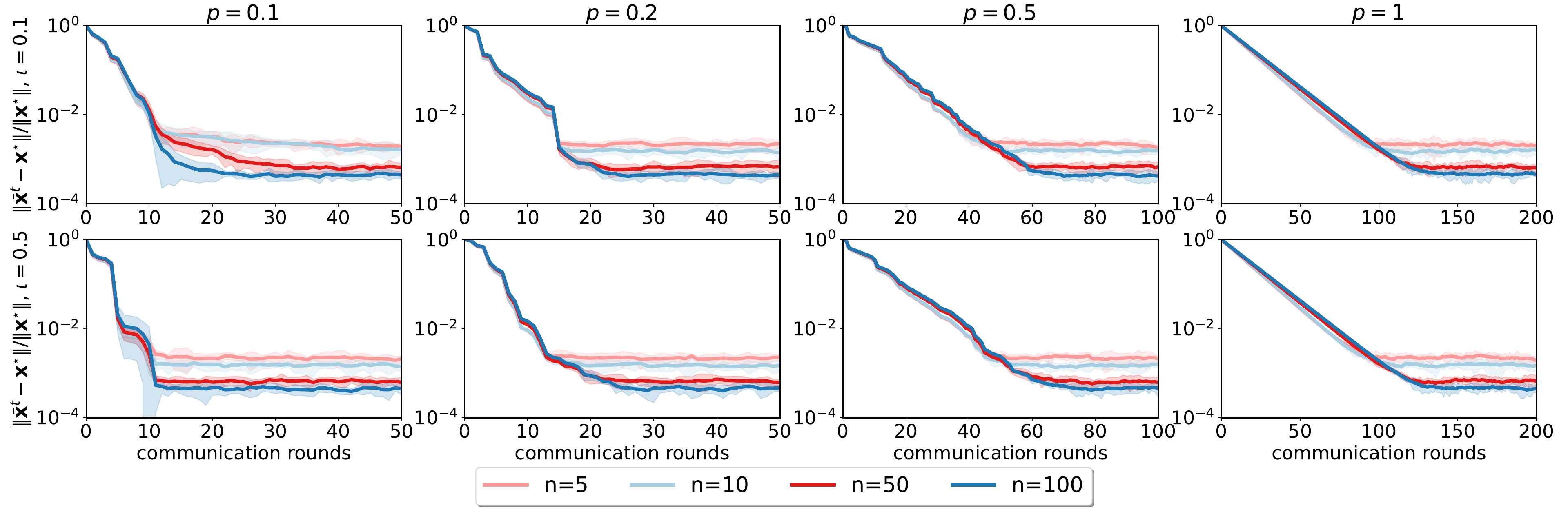}
\caption{Experimental results for ProxSkip to logistic regression problem with a strongly convex regularizer $r(\M{x})=\frac{1}{2}\|\M{x}\|^2$ over ijcnn1 dataset.}
\label{Sim:StronglyConvex-F1}
\end{figure*}

\begin{figure*}[!t]
\centering
\setlength{\abovecaptionskip}{0pt}
\includegraphics[width=\linewidth]{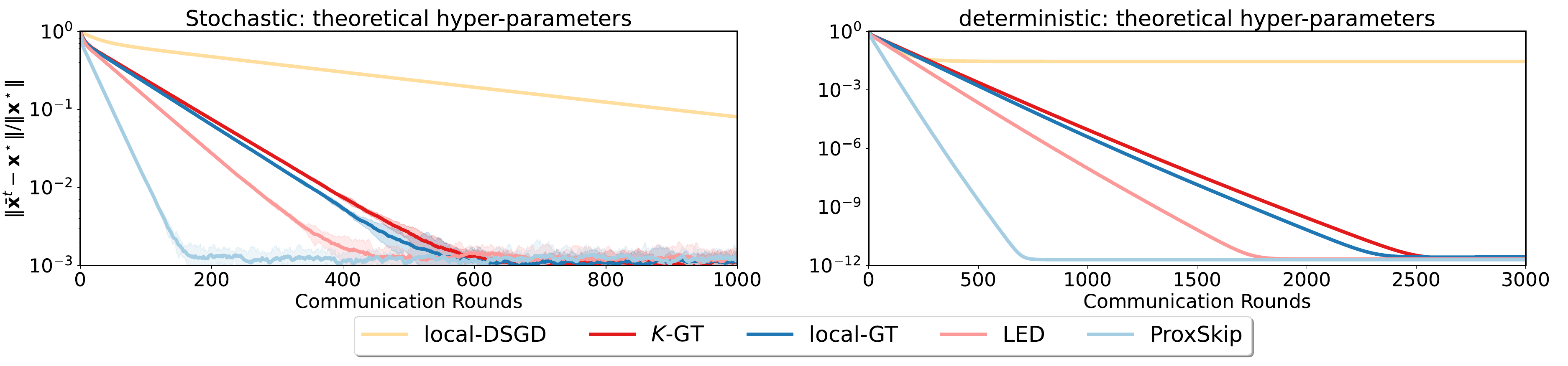}
\caption{Experimental results of logistic regression problem on the ijcnn1 dataset with regularizer $r(\M{x})=\frac{L}{200}\|\M{x}\|^2$, where $1-\lambda_2\approx0.25$. We set $p=\nicefrac{1}{\sqrt{(1-\lambda_2)\kappa}}\approx0.2$ (the choice suggested by the deterministic bound in Lemma~\ref{THM-1-PER}).}
\label{Sim:StronglyConvex-F2}
\end{figure*}

\begin{figure*}[!t]
\centering
\includegraphics[width=1\linewidth]{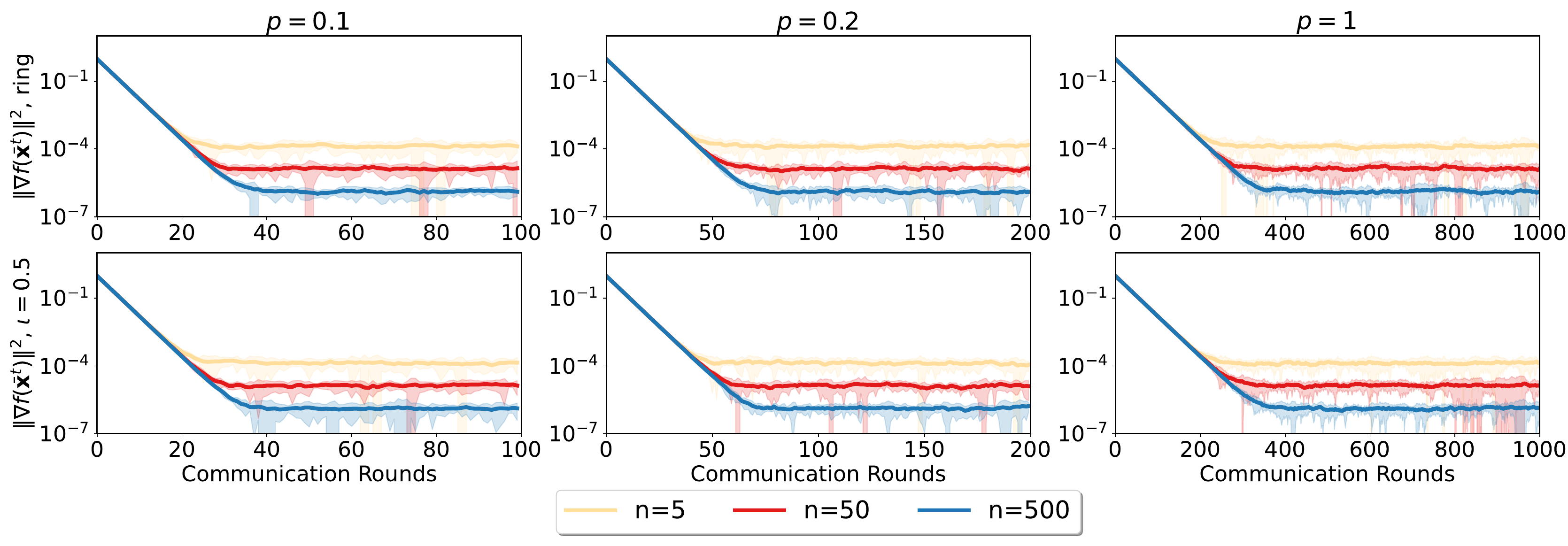}
\caption{Linear speedup of ProxSkip in non-convex settings over ijcnn1 dataset.}
\label{Sim:F3}
\end{figure*}

\begin{figure*}[!t]
\centering
\includegraphics[width=1\linewidth]{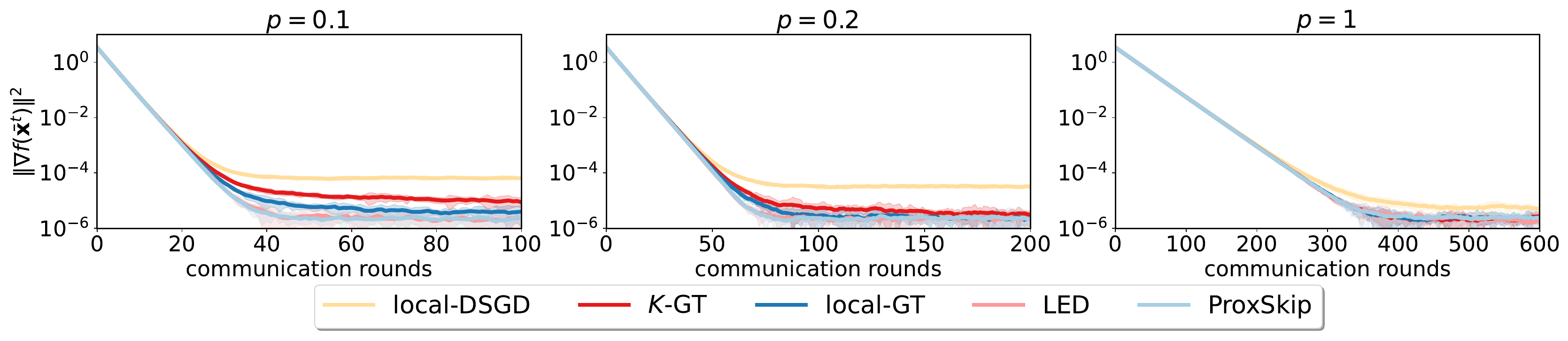}
\caption{Experimental comparison with the same stepsize in non-convex settings over ijcnn1 dataset.}
\label{Sim:F1}
\end{figure*}

\begin{figure*}[!t]
\centering
\includegraphics[width=1\linewidth]{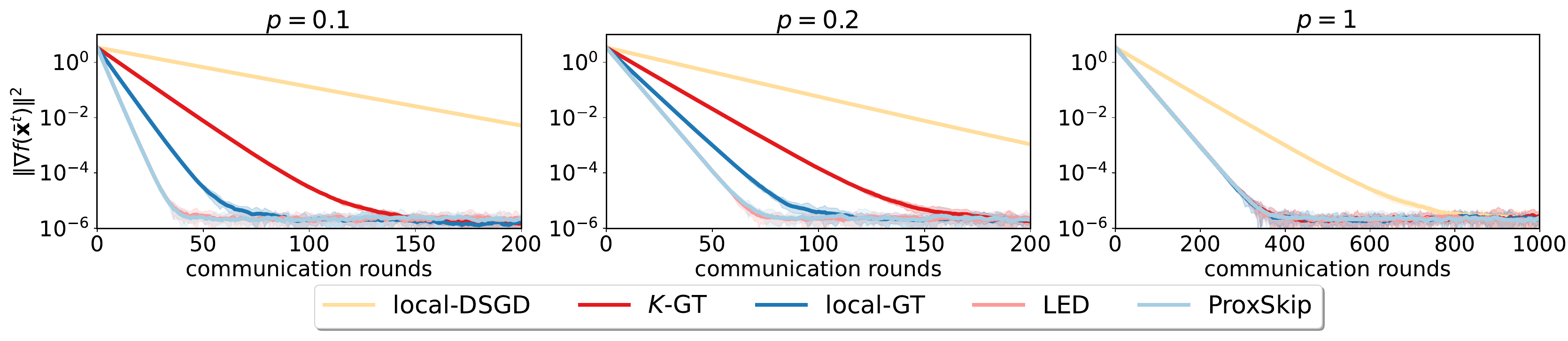}
\caption{Experimental comparison with tuning stepsize in non-convex settings over ijcnn1 dataset.}
\label{Sim:F2}
\end{figure*}

\emph{1) Strongly convex regularizer:} We choose the regularizer $r(\M{x})=\frac{1}{2}\|\M{x}\|^2$ to demonstrate the results in the strongly convex setting. The results are shown in Fig. \ref{Sim:StronglyConvex-F1}. The relative error $\nicefrac{\|\bar{\M{x}}^t-\M{x}^{\star}\|}{\|\M{x}^\star\|}$ is shown on the $y$-axis. We show the performance of ProxSkip at different number of nodes $n$, network connectivity $\iota$, and communication probability $p$. The results show that, when the number of nodes is increased, the relative errors of ProxSkip is reduced, which validates our results about linear speedup. Moreover, Fig. \ref{Sim:StronglyConvex-F1} shows that we can save on communication rounds by reducing $p$, i.e., increasing the number of local steps reduces the amount of communication required to achieve the same level of accuracy. In addition, we compare the performance of ProxSkip with Local-DSGD \cite{Wang2021,Stich2020}, local-GT \cite{HuangYan2023,Nguyen2022}, $K$-GT \cite{Liu2023}, and LED \cite{Alghunaim2023} under stochastic and deterministic settings, respectively, and the results are shown in Fig. \ref{Sim:StronglyConvex-F2}.

\emph{2) Non-convex regularizer:} We choose the regularizer $r(\M{x})=\sum_{j=1}^{d}\frac{\M{x}(j)^2}{1+\M{x}(j)^2}$, to demonstrate the results in the nonconvex setting, where $\M{x}=\mathrm{col}\{\M{x}(j)\}_{j=1}^d\in \mathbb{R}^d$. In this case, we first show that the ProxSkip algorithm achieves linear speedup even in the nonconvex setting. In Fig. \ref{Sim:F3}, we conduct experiments at the ring and at $\iota=0.5$, respectively, and the experimental results are consistent with the theoretical results of Theorem \ref{MainResult-Convergence-rate-T}. Setting $n=10$, we also compare ProxSkip to the distributed methods Local-DSGD \cite{Wang2021,Stich2020}, local-GT \cite{HuangYan2023,Nguyen2022}, $K$-GT \cite{Liu2023}, and LED \cite{Alghunaim2023} for different local steps $\nicefrac{1}{p}=10,5,1$. We use the same stepsize $\alpha=0.01$ for all algorithms. Fig. \ref{Sim:F1} and Fig. \ref{Sim:F2} show that ProxSkip and LED perform similarly and outperform the other methods as we increase the number of local steps. Moreover, from $\nicefrac{1}{p}=1$ to $\nicefrac{1}{p}=10$ in Fig. \ref{Sim:F1} and Fig. \ref{Sim:F2}, we observe that local steps improve the communication efficiency of ProxSkip.

\section{Conclusion}
This paper revisits the convergence bounds of ProxSkip for distributed stochastic optimization. We present a new analysis with a novel proof technique applicable to stochastic non-convex, convex, and strongly convex settings. This analysis gives bounds comparable to those of state-of-the-art distributed algorithms \cite{Stich2020,HuangYan2023,Nguyen2022,Liu2023,Alghunaim2023}. We establish that the leading expected communication complexity is $O(p\sigma^2/(n\epsilon^2))$ for the non-convex and convex cases, and $\tilde O(p\sigma^2/(n\epsilon))$ for the strongly convex case, indicating that ProxSkip can achieve acceleration by $p$ and $n$. Our proposed proof technique overcomes the analytical limitations of prior work \cite{ProxSkip,TAMUNA,CompressedScaffnew,LoCoDL,VR-ProxSkip,T-ProxSkip,RandProx} and might be of independent interest in the community.

\newpage
\newpage

\onecolumn
{\huge \noindent Appendix}

\section{Preliminaries}\label{SEC-PROOF}
\subsection{Basic Facts}
For any matrices $\M{A}$ and $\M{B}$ of the same size, their inner product is denoted as $\langle\M{A},\M{B}\rangle=\mathrm{Trace}(\M{A}\tr\M{B})$. For a given matrix $\M{A}$, the Frobenius norm is denoted by $\|\M{A}\|_{\mathrm F}$, while the spectral norm is denoted by $\|\M{A}\|$. We also define the following notations to simplify the analysis:
\begin{align*}
\bar{\M{x}}^t=\frac1n(\M{X}^t)\tr\M{1},\ \bar{\M{s}}^t=\frac1n(\M{S}^t)\tr\M{1},\qquad \bar{\M{X}}^t=\M{1}(\bar{\M{x}}^t)\tr,\ \overline{\nabla F}(\M{X}^t)=\frac1n[\nabla F(\M{X}^t)]\tr\M{1},
\end{align*}
where $\M{S}^t=\M{G}^t-\nabla F(\M{X}^t)$ is the gradient noise.

The pre-sampling history $\mathcal F^t$ is defined in Section~\ref{SEC-II}. After the current samples are revealed, let $\mathcal G^t=\mathcal F^t\vee\sigma(\xi^t)$; after the communication variable is revealed, $\mathcal F^{t+1}=\mathcal G^t\vee\sigma(\theta_t)$. Thus, $\mathcal F^t\subseteq\mathcal G^t\subseteq\mathcal F^{t+1}$. The variables $(\M{X}^t,\M{Y}^t)$ are $\mathcal F^t$-measurable, $(\M{G}^t,\M{S}^t,\hat{\M{Z}}^t)$ are $\mathcal G^t$-measurable, and $(\M{X}^{t+1},\M{Y}^{t+1})$ are $\mathcal F^{t+1}$-measurable. The Bernoulli variable $\theta_t$ is independent of $\mathcal G^t$.

Assumption \ref{StochasticGradient1} gives
\begin{equation}
\label{eq:noise-history}
\Exp{\M{S}^t\mid\mathcal F^t}=0,\ \Exp{\|\M{S}^t\|_{\mathrm F}^2\mid\mathcal F^t}\le n\sigma^2,\qquad \Exp{\bar{\M{s}}^t\mid\mathcal F^t}=0,\ \Exp{\|\bar{\M{s}}^t\|^2\mid\mathcal F^t}\le\frac{\sigma^2}{n}.
\end{equation}
Indeed, for $i\ne j$, conditional independence and conditional unbiasedness imply $\Exp{\langle\M{s}_i^t,\M{s}_j^t\rangle\mid\mathcal F^t}=0$. Hence, the last bound follows by summing the conditional second moments and dividing by $n^2$. {Independence of $\theta_t$ from $\mathcal G^t$ also implies $\Exp{\M{S}^t\mid\mathcal F^t\vee\sigma(\theta_t)}=0$.} Since $\M{1}\tr\M{Y}^0=0$ and $\M{W}_a$ is doubly stochastic, \eqref{Update:RandCom} gives $\M{1}\tr\M{Y}^t=0$ for every $t$. Averaging the primal update therefore yields
\begin{equation}
\label{RandCom: ErrorRecursion-F1-NC}
\bar{\M{x}}^{t+1}=\bar{\M{x}}^t-\alpha\overline{\nabla F}(\M{X}^t)-\alpha\bar{\M{s}}^t.
\end{equation}
In particular, $\bar{\M{x}}^{t+1}$ is $\mathcal G^t$-measurable and does not depend on $\theta_t$. Under Assumption \ref{MixingMatrix}, choose an orthonormal eigenbasis $\hat{\M{P}}=[\M{p}_2,\ldots,\M{p}_n]\in\mathbb R^{n\times(n-1)}$ for the non-consensus subspace. Then
\begin{align*}
\M{W}=\frac1n\M{11}\tr+\hat{\M{P}}\hat{\M{\Lambda}}\hat{\M{P}}\tr,\ \hat{\M{\Lambda}}=\mathrm{diag}(\lambda_2,\ldots,\lambda_n),
\end{align*}
where $\hat{\M{P}}\tr\hat{\M{P}}=\M{I}$, $\M{1}\tr\hat{\M{P}}=0$, and $\hat{\M{P}}\hat{\M{P}}\tr=\M{I}-\frac1n\M{11}\tr$.

\section{Proof of Theorem \ref{MainResult-Convergence-rate-T} and Corollaries \ref{MainResult-Convergence-rate-tighter-stepsize} and \ref{MainResult-Convergence-rate-C}}\label{AP:mainproof}
\subsection{Transformation and Some Descent Inequalities}
Throughout the proof, take $\chi\ge4$. For $i=2,\ldots,n$, define
\begin{equation}
\label{eq:proof-parameters}
r_i=\frac{1-\lambda_i}{2\chi},\ m=\frac{1-\lambda_2}{2\chi},\ g=\frac{pm}{10}.
\end{equation}
Thus, $0<m\le r_i\le1/4$ and $0<g\le1/40$. The stepsize condition in Theorem \ref{MainResult-Convergence-rate-T} is $\alpha L\le g/16$.

{
To make the reference-gradient increment measurable before current sampling, introduce the auxiliary sequence
\begin{equation}
\label{eq:filtered-history}
\M{a}^0=0,\ \M{a}^{t+1}=(1-g)\M{a}^t+\alpha\bar{\M{s}}^t,\ \M{z}^t=\bar{\M{x}}^t+\M{a}^t.
\end{equation}
Combining \eqref{eq:filtered-history} and \eqref{RandCom: ErrorRecursion-F1-NC}, we have
\begin{equation}
\label{eq:predictable-reference}
\M{z}^{t+1} =\bar{\M{x}}^t-\alpha\overline{\nabla F}(\M{X}^t)-\alpha\bar{\M{s}}^t +(1-g)\M{a}^t+\alpha\bar{\M{s}}^t =\M{z}^t-\alpha\overline{\nabla F}(\M{X}^t)-g\M{a}^t.
\end{equation}
By induction, $\M{a}^t$, $\M{z}^t$, and $\M{z}^{t+1}$ are $\mathcal F^t$-measurable. These auxiliary variables are used only in the analysis.
}
Here, we introduce the auxiliary variable $\M{R}^t=\M{Y}^t+\alpha\nabla F(\M{1}(\M{z}^t)\tr)$ and write
\begin{align*}
\M{\Sigma}_1^t=\nabla F(\M{X}^t)-\nabla F(\M{1}(\M{z}^t)\tr),\qquad \M{\Sigma}_2^t=\nabla F(\M{1}(\M{z}^t)\tr)-\nabla F(\M{1}(\M{z}^{t+1})\tr).
\end{align*}
{In particular, $\M{R}^t$, $\M{\Sigma}_1^t$, and $\M{\Sigma}_2^t$ are $\mathcal F^t$-measurable. We keep the current gradient noise $\M{S}^t$ separate from these terms.}
By this variable, we give the following error dynamics of ProxSkip.

\begin{lem}\label{AnotherErrorDynamicsofRandCom}\label{lem:weighted-communication}
Suppose that Assumption \ref{MixingMatrix} holds and $\chi\ge4$. There exists an invertible real matrix $\M{Q}\in\mathbb R^{2(n-1)\times2(n-1)}$ such that, with
{
\begin{equation}
\label{eq:transformed-state}
\mathcal E^t=\M{Q}^{-1} \begin{bmatrix}\hat{\M{P}}\tr\M{X}^t\\\hat{\M{P}}\tr\M{R}^t\end{bmatrix} \in\mathbb R^{2(n-1)\times d},
\end{equation}
the error recursion is
\begin{equation}
\label{eq:filtered-recursion}
\mathcal E^{t+1} ={}\M{\Gamma}_{\theta_t}\mathcal E^t -\alpha\M{Q}^{-1} \begin{bmatrix} \hat{\M{\Lambda}}_{\theta_t}\hat{\M{P}}\tr\M{\Sigma}_1^t\\ p(\M{I}-\hat{\M{\Lambda}}_{\theta_t})\hat{\M{P}}\tr\M{\Sigma}_1^t+\hat{\M{P}}\tr\M{\Sigma}_2^t \end{bmatrix} -\alpha\M{C}_{\theta_t}\M{S}^t,
\end{equation}
where
\begin{align*}
\hat{\M{\Lambda}}_\theta&=\M{I}-\theta\,\mathrm{diag}(r_2,\ldots,r_n),\qquad \M{\Gamma}_\theta=\M{Q}^{-1} \begin{bmatrix} \hat{\M{\Lambda}}_\theta&-\hat{\M{\Lambda}}_\theta\\ p(\M{I}-\hat{\M{\Lambda}}_\theta)&\M{I}-p(\M{I}-\hat{\M{\Lambda}}_\theta) \end{bmatrix}\M{Q},\\
\M{C}_\theta&=\M{Q}^{-1} \begin{bmatrix} \hat{\M{\Lambda}}_\theta\hat{\M{P}}\tr\\ p(\M{I}-\hat{\M{\Lambda}}_\theta)\hat{\M{P}}\tr \end{bmatrix}\in\mathbb R^{2(n-1)\times n}.
\end{align*}
In particular, the matrix $\M{\Gamma}_\theta$ depends on the communication variable $\theta$. For $\theta\sim\mathrm{Bernoulli}(p)$, it satisfies
\begin{equation}
\label{eq:transformed-contraction}
\mathbb E_\theta[\M{\Gamma}_\theta\tr\M{\Gamma}_\theta] \preceq(1-g)\M{I},\qquad \mathbb E_\theta[\M{C}_\theta\tr\M{C}_\theta] \preceq\hat{\M{P}}\hat{\M{P}}\tr\preceq\M{I}.
\end{equation}
}
Moreover, $\|\M{Q}\|^2\le8/5$ and $\|\M{Q}^{-1}\|^2\le5/(4p^2m)$.
\end{lem}
\begin{IEEEproof}
Since $\M{I}-\M{W}_a=\frac1{2\chi}\M{W}_b^2$, the updates \eqref{Update:RandCom} give
\begin{align*}
\M{X}^{t+1} =(1-\theta_t)\hat{\M{Z}}^t+\theta_t\M{W}_a\hat{\M{Z}}^t =\left(\M{I}-\frac{\theta_t}{2\chi}\M{W}_b^2\right)\hat{\M{Z}}^t,\qquad \M{Y}^{t+1} =\M{Y}^t+p(\hat{\M{Z}}^t-\M{X}^{t+1}) =\M{Y}^t+\frac{p\theta_t}{2\chi}\M{W}_b^2\hat{\M{Z}}^t.
\end{align*}
By the definitions of $\M{R}^t$ and $\M{\Sigma}_2^t$, we have
\begin{align*}
\M{R}^{t+1}-\M{R}^t =\M{Y}^{t+1}-\M{Y}^t+\alpha\big(\nabla F(\M{1}(\M{z}^{t+1})\tr)-\nabla F(\M{1}(\M{z}^t)\tr)\big) =\frac{p\theta_t}{2\chi}\M{W}_b^2\hat{\M{Z}}^t-\alpha\M{\Sigma}_2^t.
\end{align*}
Consequently, Algorithm \ref{RandCom} is equivalent to
\begin{align*}
\hat{\M{Z}}^t=\M{X}^t-\M{R}^t-\alpha\M{\Sigma}_1^t-\alpha\M{S}^t,\qquad \M{X}^{t+1}=\left(\M{I}-\frac{\theta_t}{2\chi}\M{W}_b^2\right)\hat{\M{Z}}^t,\qquad \M{R}^{t+1}=\M{R}^t+\frac{p\theta_t}{2\chi}\M{W}_b^2\hat{\M{Z}}^t-\alpha\M{\Sigma}_2^t.
\end{align*}
Multiplying the last two identities by $\hat{\M{P}}\tr$ and using $\hat{\M{P}}\tr\M{W}_b^2=(\M{I}-\hat{\M{\Lambda}})\hat{\M{P}}\tr$, we obtain
\begin{align*}
\hat{\M{P}}\tr\M{X}^{t+1} =\hat{\M{\Lambda}}_{\theta_t}\hat{\M{P}}\tr(\M{X}^t-\M{R}^t-\alpha\M{\Sigma}_1^t-\alpha\M{S}^t),\qquad \hat{\M{P}}\tr\M{R}^{t+1} =\hat{\M{P}}\tr\M{R}^t-\alpha\hat{\M{P}}\tr\M{\Sigma}_2^t +p(\M{I}-\hat{\M{\Lambda}}_{\theta_t})\hat{\M{P}}\tr(\M{X}^t-\M{R}^t-\alpha\M{\Sigma}_1^t-\alpha\M{S}^t).
\end{align*}
Stacking these identities and applying $\M{Q}^{-1}$ gives \eqref{eq:filtered-recursion}.

We now construct $\M{Q}$ and establish its bounds. Since all blocks of the transition matrix are diagonal, there exists a permutation matrix $\M{Q}_1$ such that its conjugation by $\M{Q}_1$ is $\mathrm{blkdiag}\{\M{H}_{\theta,i}\}_{i=2}^n$, where
\begin{align*}
\M{H}_{\theta,i}=\begin{bmatrix} 1-\theta r_i&-(1-\theta r_i)\\ p\theta r_i&1-p\theta r_i \end{bmatrix}.
\end{align*}
{For each mode, define the real symmetric matrix
\begin{equation}
\label{eq:modal-metric}
\M{K}_i=\begin{bmatrix} 1&\dfrac{1-p}{4p}\\[2mm] \dfrac{1-p}{4p}&\dfrac{1+r_i(1-3p-2p^2)/4}{p^2r_i} \end{bmatrix}.
\end{equation}
We will show that
\begin{equation}
\label{eq:metric-comparison}
\frac58\begin{bmatrix}1&0\\0&(p^2r_i)^{-1}\end{bmatrix} \preceq\M{K}_i\preceq \frac54\begin{bmatrix}1&0\\0&(p^2r_i)^{-1}\end{bmatrix}.
\end{equation}
Thus, $\M{K}_i$ is positive definite, and we may set
\begin{equation}
\label{eq:Q-construction}
\M{Q}=\M{Q}_1\tr\mathrm{blkdiag}\{\M{K}_i^{-1/2}\}_{i=2}^n,\qquad \M{Q}^{-1}=\mathrm{blkdiag}\{\M{K}_i^{1/2}\}_{i=2}^n\M{Q}_1.
\end{equation}
}
Fix a mode in \eqref{eq:modal-metric} and omit the index $i$. Set $\M{J}=\mathrm{diag}(1,p\sqrt r)$. Direct multiplication gives
\begin{align*}
\M{JKJ}=\begin{bmatrix} 1&(1-p)\sqrt r/4\\ (1-p)\sqrt r/4&1+r(1-3p-2p^2)/4 \end{bmatrix}.
\end{align*}
Its diagonal entries lie in $[3/4,17/16]$, and its off-diagonal entry has magnitude at most $1/8$. Hence $\frac58\M{I}\preceq\M{JKJ}\preceq\frac54\M{I}$, proving \eqref{eq:metric-comparison}. Since $p^2r_i\le1$ and $r_i\ge m$, it follows that
\begin{align*}
\|\M{Q}\|^2=\max_i\|\M{K}_i^{-1}\|\le\frac85,\qquad \|\M{Q}^{-1}\|^2=\max_i\|\M{K}_i\|\le\frac5{4p^2m}.
\end{align*}
Let $c=(1-p)(2p+3)\in[0,3]$ and $\M{B}=\M{K}-\mathbb E_\theta[\M{H}_\theta\tr\M{K}\M{H}_\theta]$. Direct multiplication gives
\begin{align*}
B_{11}=pr\left(\frac{1+p}{2}-\frac{cr}{4}\right),\qquad B_{12}=B_{21}=\frac{pcr^2}{4},\qquad B_{22}=\frac{1+p}{2p}-\frac{p(1+p)r}{2}-\frac{pcr^2}{4}.
\end{align*}
The diagonal entries of $\M{JBJ}/(pr)$ are at least $5/16$ and $21/64$, respectively, and its off-diagonal entry has magnitude at most $3/32$. Therefore,
$\frac{\M{JBJ}}{pr}\succeq\frac7{32}\M{I}\succeq\frac1{10}\M{JKJ}$,
which implies
\begin{equation}
\label{eq:metric-contraction}
\mathbb E_\theta[\M{H}_{\theta,i}\tr\M{K}_i\M{H}_{\theta,i}] \preceq\left(1-\frac{pr_i}{10}\right)\M{K}_i.
\end{equation}
By \eqref{eq:Q-construction}, the blocks of $\M{\Gamma}_\theta$ are $\M{K}_i^{1/2}\M{H}_{\theta,i}\M{K}_i^{-1/2}$. Congruence of \eqref{eq:metric-contraction} by $\M{K}_i^{-1/2}$ and $pr_i/10\ge g$ prove the first inequality in \eqref{eq:transformed-contraction}. {To bound the complete noise map, write $\M{e}_1=(1,0)\tr$. Since $K_{11}=1$, the preceding expression for $B_{11}$ yields
\begin{equation}
\label{eq:noise-gain}
\mathbb E_\theta[(\M{H}_{\theta,i}\M{e}_1)\tr\M{K}_i(\M{H}_{\theta,i}\M{e}_1)] =1-\frac{pr_i(1+p)}2+\frac{pr_i^2(1-p)(2p+3)}4\le1.
\end{equation}
Here $B_{11}\ge5pr_i/16>0$. The $i$th two-row block of $\M{C}_\theta$ is $\M{K}_i^{1/2}\M{H}_{\theta,i}\M{e}_1\M{p}_i\tr$. Consequently, for any $\M{A}\in\mathbb R^{n\times d}$, \eqref{eq:noise-gain} gives
\begin{align*}
\mathbb E_\theta\|\M{C}_\theta\M{A}\|_{\mathrm F}^2 =\sum_{i=2}^n\mathbb E_\theta\big[(\M{H}_{\theta,i}\M{e}_1)\tr\M{K}_i(\M{H}_{\theta,i}\M{e}_1)\big]\|\M{p}_i\tr\M{A}\|^2 \le\sum_{i=2}^n\|\M{p}_i\tr\M{A}\|^2 =\|\hat{\M{P}}\tr\M{A}\|_{\mathrm F}^2\le\|\M{A}\|_{\mathrm F}^2.
\end{align*}
This proves the second inequality in \eqref{eq:transformed-contraction}; no matrix factor has been omitted from the noise norm.
}
\end{IEEEproof}

Define the error energy
\begin{equation}
\label{eq:joint-energy}
U^t=\frac1n\|\mathcal E^t\|_{\mathrm F}^2+\frac{16\alpha^2L^2}{g^2}\|\M{a}^t\|^2.
\end{equation}
Since $\hat{\M{P}}\hat{\M{P}}\tr=\M{I}-\frac1n\M{11}\tr$, \eqref{eq:transformed-state} and $\|\M{Q}\|^2\le8/5$ give
\begin{equation}
\label{UsefulInequality1}
\frac1n\|\M{X}^t-\bar{\M{X}}^t\|_{\mathrm F}^2 =\frac1n\|\hat{\M{P}}\tr\M{X}^t\|_{\mathrm F}^2 \le\frac{\|\M{Q}\|^2}{n}\|\mathcal E^t\|_{\mathrm F}^2 \le\frac8{5n}\|\mathcal E^t\|_{\mathrm F}^2\le\frac85 U^t.
\end{equation}
Based on the transformation in Lemma \ref{AnotherErrorDynamicsofRandCom}, we give the following descent inequalities.

\begin{lem}\label{NCVX-Lemma-DescentInequalities}\label{lem:filtered-consensus}
Suppose that Assumptions \ref{MixingMatrix}, \ref{ASS1}, and \ref{StochasticGradient1} hold. With $\chi\ge4$ and $0<\alpha L\le g/16$, we have
\begin{align}
\label{NCVX-Lemma-DescentInequalities1} \Exp{f(\bar{\M{x}}^{t+1})\mid\mathcal F^t}&\le f(\bar{\M{x}}^t)-\frac\alpha2\|\nabla f(\bar{\M{x}}^t)\|^2+\frac{L\alpha^2\sigma^2}{2n} -\frac{\alpha(1-\alpha L)}2\|\overline{\nabla F}(\M{X}^t)\|^2 +\frac{\alpha L^2}{2n}\|\M{X}^t-\bar{\M{X}}^t\|_{\mathrm F}^2,\\
\Exp{U^{t+1}\mid\mathcal F^t}&\le\left(1-\frac g2\right)U^t+2\alpha^2\sigma^2 +\frac{25\alpha^4L^2}{p^2mg}\|\overline{\nabla F}(\M{X}^t)\|^2.\label{eq:joint-consensus}
\end{align}
For the initialization in Algorithm \ref{RandCom},
\begin{equation}
\label{eq:energy-initialization}
U^0=\frac1n\|\mathcal E^0\|_{\mathrm F}^2\le\frac{5\alpha^2\varsigma_0^2}{4p^2m}.
\end{equation}
\end{lem}
\begin{IEEEproof}
\textbf{\emph{Proof of the inequality \eqref{NCVX-Lemma-DescentInequalities1}.}}
Since $f$ is $L$-smooth, it holds that
\begin{align*}
f(\bar{\M{x}}^{t+1}) \le{}f(\bar{\M{x}}^t)+\langle\nabla f(\bar{\M{x}}^t),\bar{\M{x}}^{t+1}-\bar{\M{x}}^t\rangle +\frac L2\|\bar{\M{x}}^{t+1}-\bar{\M{x}}^t\|^2.
\end{align*}
From \eqref{RandCom: ErrorRecursion-F1-NC}, we have
\begin{align*}
f(\bar{\M{x}}^{t+1}) \le{}f(\bar{\M{x}}^t)-\alpha\langle\nabla f(\bar{\M{x}}^t),\overline{\nabla F}(\M{X}^t)+\bar{\M{s}}^t\rangle +\frac{L\alpha^2}{2}\|\overline{\nabla F}(\M{X}^t)+\bar{\M{s}}^t\|^2.
\end{align*}
Taking the conditional expectation given $\mathcal F^t$ and using \eqref{eq:noise-history}, we obtain
\begin{align*}
\Exp{f(\bar{\M{x}}^{t+1})\mid\mathcal F^t} &\le f(\bar{\M{x}}^t)-\alpha\langle\nabla f(\bar{\M{x}}^t),\overline{\nabla F}(\M{X}^t)\rangle +\frac{L\alpha^2}{2}\left(\|\overline{\nabla F}(\M{X}^t)\|^2 +\Exp{\|\bar{\M{s}}^t\|^2\mid\mathcal F^t}\right)\\
&\le f(\bar{\M{x}}^t)-\alpha\langle\nabla f(\bar{\M{x}}^t),\overline{\nabla F}(\M{X}^t)\rangle +\frac{L\alpha^2}{2}\|\overline{\nabla F}(\M{X}^t)\|^2+\frac{L\alpha^2\sigma^2}{2n}.
\end{align*}
Since $2\langle a,b\rangle=\|a\|^2+\|b\|^2-\|a-b\|^2$, it follows that
\begin{align*}
-\alpha\langle\nabla f(\bar{\M{x}}^t),\overline{\nabla F}(\M{X}^t)\rangle =-\frac\alpha2\|\nabla f(\bar{\M{x}}^t)\|^2-\frac\alpha2\|\overline{\nabla F}(\M{X}^t)\|^2 +\frac\alpha2\|\overline{\nabla F}(\M{X}^t)-\nabla f(\bar{\M{x}}^t)\|^2.
\end{align*}
Combining these inequalities gives
\begin{align*}
\Exp{f(\bar{\M{x}}^{t+1})\mid\mathcal F^t} \le f(\bar{\M{x}}^t)-\frac\alpha2\|\nabla f(\bar{\M{x}}^t)\|^2 -\frac{\alpha(1-\alpha L)}2\|\overline{\nabla F}(\M{X}^t)\|^2 +\frac\alpha2\|\overline{\nabla F}(\M{X}^t)-\nabla f(\bar{\M{x}}^t)\|^2 +\frac{L\alpha^2\sigma^2}{2n}.
\end{align*}
Moreover, Jensen's inequality and $L$-smoothness imply
\begin{align*}
\|\overline{\nabla F}(\M{X}^t)-\nabla f(\bar{\M{x}}^t)\|^2 =\left\|\frac1n\sum_{i=1}^n\big(\nabla f_i(\M{x}_i^t)-\nabla f_i(\bar{\M{x}}^t)\big)\right\|^2 \le\frac1n\sum_{i=1}^n\|\nabla f_i(\M{x}_i^t)-\nabla f_i(\bar{\M{x}}^t)\|^2 \le\frac{L^2}{n}\|\M{X}^t-\bar{\M{X}}^t\|_{\mathrm F}^2.
\end{align*}
Thus, the descent inequality \eqref{NCVX-Lemma-DescentInequalities1} holds. We retain its negative $\|\overline{\nabla F}(\M{X}^t)\|^2$ term for the subsequent Lyapunov estimate.

\textbf{\emph{Proof of the inequality \eqref{eq:joint-consensus}.}}
Writing
\begin{align*}
\mathbb G^t =\M{\Gamma}_{\theta_t}\mathcal E^t-\alpha\M{C}_{\theta_t}\M{\Sigma}_1^t -\alpha\M{Q}^{-1}\begin{bmatrix}0\\\hat{\M{P}}\tr\M{\Sigma}_2^t\end{bmatrix},
\end{align*}
we have $\mathcal E^{t+1}=\mathbb G^t-\alpha\M{C}_{\theta_t}\M{S}^t$ by \eqref{eq:filtered-recursion}.
{The variables $\mathbb G^t$ and $\M{C}_{\theta_t}$ are $\mathcal F^t\vee\sigma(\theta_t)$-measurable. Independence of $\theta_t$ from $\mathcal G^t$ and \eqref{eq:noise-history} give
\begin{align*}
\Exp{\langle\mathbb G^t,\M{C}_{\theta_t}\M{S}^t\rangle\mid\mathcal F^t\vee\sigma(\theta_t)} =\left\langle\mathbb G^t,\M{C}_{\theta_t}\Exp{\M{S}^t\mid\mathcal F^t\vee\sigma(\theta_t)}\right\rangle=0.
\end{align*}
Expanding the squared norm under this conditioning and then applying the tower property, we obtain
\begin{align*}
\Exp{\|\mathcal E^{t+1}\|_{\mathrm F}^2\mid\mathcal F^t} =\Exp{\|\mathbb G^t\|_{\mathrm F}^2\mid\mathcal F^t} +\alpha^2\Exp{\|\M{C}_{\theta_t}\M{S}^t\|_{\mathrm F}^2\mid\mathcal F^t}.
\end{align*}
To bound the noise term, first condition on $\mathcal G^t$, with respect to which $\M{S}^t$ is measurable. By \eqref{eq:transformed-contraction} and independence of $\theta_t$ from $\mathcal G^t$,
\begin{align*}
\Exp{\|\M{C}_{\theta_t}\M{S}^t\|_{\mathrm F}^2\mid\mathcal G^t} =\mathrm{Trace}\!\left((\M{S}^t)\tr\mathbb E_\theta[\M{C}_\theta\tr\M{C}_\theta]\M{S}^t\right) \le\|\hat{\M{P}}\tr\M{S}^t\|_{\mathrm F}^2.
\end{align*}
Applying the tower property with $\mathcal F^t\subseteq\mathcal G^t$ yields
\begin{align*}
\alpha^2\Exp{\|\M{C}_{\theta_t}\M{S}^t\|_{\mathrm F}^2\mid\mathcal F^t} \le\alpha^2\Exp{\|\hat{\M{P}}\tr\M{S}^t\|_{\mathrm F}^2\mid\mathcal F^t} \le n\alpha^2\sigma^2.
\end{align*}
}
We next bound $\Exp{\|\mathbb G^t\|_{\mathrm F}^2\mid\mathcal F^t}$. Using $\|a+b\|^2\le(1+g/4)\|a\|^2+(1+4/g)\|b\|^2$ and then $\|a+b\|^2\le2\|a\|^2+2\|b\|^2$, we have
\begin{align*}
\|\mathbb G^t\|_{\mathrm F}^2 \le{}\left(1+\frac g4\right)\|\M{\Gamma}_{\theta_t}\mathcal E^t\|_{\mathrm F}^2+2\alpha^2\left(1+\frac4g\right)\|\M{C}_{\theta_t}\M{\Sigma}_1^t\|_{\mathrm F}^2 +2\alpha^2\left(1+\frac4g\right)\left\|\M{Q}^{-1}\begin{bmatrix}0\\\hat{\M{P}}\tr\M{\Sigma}_2^t\end{bmatrix}\right\|_{\mathrm F}^2.
\end{align*}
Since $\mathcal E^t$ and $\M{\Sigma}_1^t$ are $\mathcal F^t$-measurable, \eqref{eq:transformed-contraction} implies
\begin{align*}
\Exp{\|\M{\Gamma}_{\theta_t}\mathcal E^t\|_{\mathrm F}^2\mid\mathcal F^t} \le(1-g)\|\mathcal E^t\|_{\mathrm F}^2,\qquad \Exp{\|\M{C}_{\theta_t}\M{\Sigma}_1^t\|_{\mathrm F}^2\mid\mathcal F^t} \le\|\M{\Sigma}_1^t\|_{\mathrm F}^2.
\end{align*}
Also, $\|\M{Q}^{-1}\|^2\le5/(4p^2m)$ gives
\begin{align*}
\left\|\M{Q}^{-1}\begin{bmatrix}0\\\hat{\M{P}}\tr\M{\Sigma}_2^t\end{bmatrix}\right\|_{\mathrm F}^2 \le\frac5{4p^2m}\|\M{\Sigma}_2^t\|_{\mathrm F}^2.
\end{align*}
Combining these bounds, $(1+g/4)(1-g)\le1-3g/4$, and $2+8/g\le10/g$, we obtain
\begin{equation}
\label{eq:raw-filtered-energy}
\frac1n\Exp{\|\mathcal E^{t+1}\|_{\mathrm F}^2\mid\mathcal F^t} \le{}\frac{1-3g/4}{n}\|\mathcal E^t\|_{\mathrm F}^2 +\frac{10\alpha^2}{ng}\|\M{\Sigma}_1^t\|_{\mathrm F}^2 +\frac{25\alpha^2}{2np^2mg}\|\M{\Sigma}_2^t\|_{\mathrm F}^2+\alpha^2\sigma^2.
\end{equation}

We then bound the two gradient-difference terms. Since $\M{z}^t=\bar{\M{x}}^t+\M{a}^t$ and $\M{1}\tr(\M{X}^t-\bar{\M{X}}^t)=0$, the consensus and average components are orthogonal. Thus,
\begin{align*}
\|\M{\Sigma}_1^t\|_{\mathrm F}^2 \le L^2\|\M{X}^t-\M{1}(\M{z}^t)\tr\|_{\mathrm F}^2 =L^2\|\M{X}^t-\bar{\M{X}}^t-\M{1}(\M{a}^t)\tr\|_{\mathrm F}^2 =L^2\|\M{X}^t-\bar{\M{X}}^t\|_{\mathrm F}^2+nL^2\|\M{a}^t\|^2.
\end{align*}
By $L$-smoothness and \eqref{eq:predictable-reference}, we also have
\begin{align*}
\|\M{\Sigma}_2^t\|_{\mathrm F}^2 \le nL^2\|\M{z}^t-\M{z}^{t+1}\|^2 =nL^2\|\alpha\overline{\nabla F}(\M{X}^t)+g\M{a}^t\|^2 \le2n\alpha^2L^2\|\overline{\nabla F}(\M{X}^t)\|^2+2ng^2L^2\|\M{a}^t\|^2.
\end{align*}
Substituting these inequalities and \eqref{UsefulInequality1} into \eqref{eq:raw-filtered-energy} yields
\begin{align*}
\frac1n\Exp{\|\mathcal E^{t+1}\|_{\mathrm F}^2\mid\mathcal F^t} \le\left(1-\frac{3g}{4}+\frac{16\alpha^2L^2}{g}\right)\frac{\|\mathcal E^t\|_{\mathrm F}^2}{n} +\frac{25\alpha^4L^2}{p^2mg}\|\overline{\nabla F}(\M{X}^t)\|^2 +\left(\frac{10\alpha^2L^2}{g}+\frac{25\alpha^2L^2g}{p^2m}\right)\|\M{a}^t\|^2+\alpha^2\sigma^2.
\end{align*}
Since $\alpha L\le g/16$ and $g^2/(p^2m)=m/100\le1/400$, we have
\begin{align*}
1-\frac{3g}{4}+\frac{16\alpha^2L^2}{g}\le1-\frac g2, \frac{10\alpha^2L^2}{g}+\frac{25\alpha^2L^2g}{p^2m}\le\frac{11\alpha^2L^2}{g}.
\end{align*}
Consequently,
\begin{equation}
\label{eq:filtered-energy-with-average}
\frac1n\Exp{\|\mathcal E^{t+1}\|_{\mathrm F}^2\mid\mathcal F^t} \le{}\frac{1-g/2}{n}\|\mathcal E^t\|_{\mathrm F}^2 +\frac{25\alpha^4L^2}{p^2mg}\|\overline{\nabla F}(\M{X}^t)\|^2 +\frac{11\alpha^2L^2}{g}\|\M{a}^t\|^2+\alpha^2\sigma^2.
\end{equation}
{Finally, $\M{a}^t$ is $\mathcal F^t$-measurable, so \eqref{eq:filtered-history} and \eqref{eq:noise-history} give
\begin{equation}
\label{eq:filter-energy}
\Exp{\|\M{a}^{t+1}\|^2\mid\mathcal F^t} =(1-g)^2\|\M{a}^t\|^2+2\alpha(1-g)\langle\M{a}^t,\Exp{\bar{\M{s}}^t\mid\mathcal F^t}\rangle +\alpha^2\Exp{\|\bar{\M{s}}^t\|^2\mid\mathcal F^t} \le(1-g)^2\|\M{a}^t\|^2+\frac{\alpha^2\sigma^2}{n}.
\end{equation}
}
Multiply \eqref{eq:filter-energy} by $16\alpha^2L^2/g^2$ and add it to \eqref{eq:filtered-energy-with-average}. By \eqref{eq:joint-energy},
\begin{align*}
\Exp{U^{t+1}\mid\mathcal F^t} \le\frac{1-g/2}{n}\|\mathcal E^t\|_{\mathrm F}^2 +\frac{25\alpha^4L^2}{p^2mg}\|\overline{\nabla F}(\M{X}^t)\|^2 +\left(\frac{11\alpha^2L^2}{g}+\frac{16\alpha^2L^2}{g^2}(1-g)^2\right)\|\M{a}^t\|^2 +\alpha^2\sigma^2+\frac{16\alpha^4L^2\sigma^2}{ng^2}.
\end{align*}
The inequalities
\begin{align*}
\frac{11\alpha^2L^2}{g}+\frac{16\alpha^2L^2}{g^2}(1-g)^2 \le\frac{16\alpha^2L^2}{g^2}\left(1-\frac g2\right),\qquad \alpha^2\sigma^2+\frac{16\alpha^4L^2\sigma^2}{ng^2} \le2\alpha^2\sigma^2
\end{align*}
therefore prove \eqref{eq:joint-consensus}.

For the initialization, $\M{a}^0=0$, $\hat{\M{P}}\tr\M{X}^0=0$, and $\M{R}^0=\alpha\nabla F(\M{X}^0)$. It follows from \eqref{eq:transformed-state} and the bound on $\|\M{Q}^{-1}\|$ that
\begin{align*}
U^0=\frac1n\|\mathcal E^0\|_{\mathrm F}^2 =\frac{\alpha^2}{n}\left\|\M{Q}^{-1}\begin{bmatrix}0\\\hat{\M{P}}\tr\nabla F(\M{X}^0)\end{bmatrix}\right\|_{\mathrm F}^2 \le\frac{5\alpha^2}{4np^2m}\|\hat{\M{P}}\tr\nabla F(\M{X}^0)\|_{\mathrm F}^2 =\frac{5\alpha^2\varsigma_0^2}{4p^2m},
\end{align*}
which proves \eqref{eq:energy-initialization}.
\end{IEEEproof}

\subsection{Convergence Analysis: Non-convex}\label{AP:nonconvex}
\begin{IEEEproof}
The stepsize bound $\alpha L\le g/16$ and \eqref{eq:proof-parameters} imply
\begin{equation}
\label{eq:forcing-absorption}
\frac{100\alpha^4L^4}{p^2mg^2} \le\frac{m}{16^4}\le\frac14.
\end{equation}
This controls the reference-gradient term in \eqref{eq:joint-consensus} using the negative squared-gradient terms of the average iteration. Define the Lyapunov function
\begin{align*}
\mathcal L^t=f(\bar{\M{x}}^t)-f^\star+\frac{2\alpha L^2}{g}U^t.
\end{align*}
By \eqref{UsefulInequality1}, the last term in \eqref{NCVX-Lemma-DescentInequalities1} satisfies
\begin{align*}
\frac{\alpha L^2}{2n}\|\M{X}^t-\bar{\M{X}}^t\|_{\mathrm F}^2 \leq\frac{4\alpha L^2}{5}U^t.
\end{align*}
Then, by \eqref{NCVX-Lemma-DescentInequalities1} and \eqref{eq:joint-consensus}, it gives
\begin{align*}
\Exp{\mathcal L^{t+1}\mid\mathcal F^t} &\leq f(\bar{\M{x}}^t)-f^\star-\frac\alpha2\|\nabla f(\bar{\M{x}}^t)\|^2 -\frac{\alpha(1-\alpha L)}2\|\overline{\nabla F}(\M{X}^t)\|^2 +\frac{4\alpha L^2}{5}U^t+\frac{L\alpha^2\sigma^2}{2n}\\
&\quad+\frac{2\alpha L^2}{g}\bigg[\left(1-\frac g2\right)U^t+2\alpha^2\sigma^2+\frac{25\alpha^4L^2}{p^2mg}\|\overline{\nabla F}(\M{X}^t)\|^2\bigg]\\
&=\mathcal L^t-\frac\alpha2\|\nabla f(\bar{\M{x}}^t)\|^2-\frac{\alpha L^2}{5}U^t+\frac{L\alpha^2\sigma^2}{2n}+\frac{4\alpha^3L^2\sigma^2}{g} -\frac\alpha2\left(1-\alpha L-\frac{100\alpha^4L^4}{p^2mg^2}\right) \|\overline{\nabla F}(\M{X}^t)\|^2.
\end{align*}
The stepsize condition gives $\alpha L\leq g/16\leq1/4$. Combining this with \eqref{eq:forcing-absorption}, we have
$1-\alpha L-\frac{100\alpha^4L^4}{p^2mg^2}\geq\frac12$.
Since $U^t\geq0$, it follows that
\begin{align*}
\Exp{\mathcal L^{t+1}\mid\mathcal F^t} \leq{}\mathcal L^t-\frac\alpha2\|\nabla f(\bar{\M{x}}^t)\|^2+\frac{L\alpha^2\sigma^2}{2n}+\frac{4\alpha^3L^2\sigma^2}{g}.
\end{align*}
Taking full expectation and summing it over $t=0,1,\ldots,T-1$, we obtain
\begin{align*}
\frac\alpha2\sum_{t=0}^{T-1}\Exp{\|\nabla f(\bar{\M{x}}^t)\|^2} \leq\mathcal L^0-\Exp{\mathcal L^T}+\frac{TL\alpha^2\sigma^2}{2n}+\frac{4T\alpha^3L^2\sigma^2}{g} \leq\mathcal L^0+\frac{TL\alpha^2\sigma^2}{2n}+\frac{4T\alpha^3L^2\sigma^2}{g},
\end{align*}
where the last inequality uses $\mathcal L^T\geq0$. Dividing both sides by $\alpha T/2$ gives
\begin{align*}
\frac1T\sum_{t=0}^{T-1}\Exp{\|\nabla f(\bar{\M{x}}^t)\|^2} \leq\frac{2\mathcal L^0}{\alpha T}+\frac{L\alpha\sigma^2}{n}+\frac{8\alpha^2L^2\sigma^2}{g}.
\end{align*}
Since $\M{X}^0=\M{1}(\M{x}^0)\tr$ and $\M{a}^0=0$, the initialization bound \eqref{eq:energy-initialization} implies
\begin{align*}
\mathcal L^0 =f(\bar{\M{x}}^0)-f^\star+\frac{2\alpha L^2}{g}U^0 \leq F_0+\frac{2\alpha L^2}{g}\frac{5\alpha^2\varsigma_0^2}{4p^2m} =F_0+\frac{100\chi^2\alpha^3L^2\varsigma_0^2}{p^3(1-\lambda_2)^2},
\end{align*}
where the equality follows from $m=(1-\lambda_2)/(2\chi)$ and $g=pm/10$. Substituting this bound and $1/g=20\chi/[p(1-\lambda_2)]$ into the preceding inequality, we have
\begin{align*}
\frac1T\sum_{t=0}^{T-1}\Exp{\|\nabla f(\bar{\M{x}}^t)\|^2} \leq\frac{2F_0}{\alpha T} +\frac{200\chi^2\alpha^2L^2\varsigma_0^2}{p^3(1-\lambda_2)^2T} +\frac{L\alpha\sigma^2}{n} +\frac{160\chi\alpha^2L^2\sigma^2}{p(1-\lambda_2)}.
\end{align*}
Thus, \eqref{NCVXMainResult1-T} holds.
\end{IEEEproof}

\subsection{Convergence Analysis: Convex}\label{AP:convex}
\begin{IEEEproof}
Smooth convexity of $f_i$ gives
\begin{align*}
\langle\nabla f_i(\M{x}_i^t),\M{x}_i^t-\M{x}^\star\rangle \ge f_i(\M{x}_i^t)-f_i(\M{x}^\star) +\frac1{2L}\|\nabla f_i(\M{x}_i^t)-\nabla f_i(\M{x}^\star)\|^2.
\end{align*}
For completeness, this inequality follows by applying the smoothness descent inequality to the nonnegative convex function $f_i(x)-f_i(\M{x}^\star)-\langle\nabla f_i(\M{x}^\star),x-\M{x}^\star\rangle$ and then interchanging the two points. Also,
\begin{align*}
f_i(\bar{\M{x}}^t) \le f_i(\M{x}_i^t)+\langle\nabla f_i(\M{x}_i^t),\bar{\M{x}}^t-\M{x}_i^t\rangle +\frac L2\|\bar{\M{x}}^t-\M{x}_i^t\|^2.
\end{align*}
Adding these inequalities and averaging over $i$, we have
\begin{align*}
\langle\bar{\M{x}}^t-\M{x}^\star,\overline{\nabla F}(\M{X}^t)\rangle \ge f(\bar{\M{x}}^t)-f^\star -\frac L{2n}\|\M{X}^t-\bar{\M{X}}^t\|_{\mathrm F}^2 +\frac1{2Ln}\sum_{i=1}^n\|\nabla f_i(\M{x}_i^t)-\nabla f_i(\M{x}^\star)\|^2.
\end{align*}
Since $\sum_i\nabla f_i(\M{x}^\star)=0$, Jensen's inequality gives
\begin{align*}
\frac1n\sum_{i=1}^n\|\nabla f_i(\M{x}_i^t)-\nabla f_i(\M{x}^\star)\|^2 \ge\left\|\frac1n\sum_{i=1}^n\big(\nabla f_i(\M{x}_i^t)-\nabla f_i(\M{x}^\star)\big)\right\|^2 =\|\overline{\nabla F}(\M{X}^t)\|^2.
\end{align*}
Combining these bounds yields
\begin{equation}
\label{eq:convex-average-inner}
\langle\bar{\M{x}}^t-\M{x}^\star,\overline{\nabla F}(\M{X}^t)\rangle \ge f(\bar{\M{x}}^t)-f^\star -\frac L{2n}\|\M{X}^t-\bar{\M{X}}^t\|_{\mathrm F}^2 +\frac1{2L}\|\overline{\nabla F}(\M{X}^t)\|^2.
\end{equation}
Expanding the squared norm in \eqref{RandCom: ErrorRecursion-F1-NC} and taking the conditional expectation given $\mathcal F^t$, we obtain
\begin{align*}
\Exp{\|\bar{\M{x}}^{t+1}-\M{x}^\star\|^2\mid\mathcal F^t} =\|\bar{\M{x}}^t-\M{x}^\star\|^2 -2\alpha\langle\bar{\M{x}}^t-\M{x}^\star,\overline{\nabla F}(\M{X}^t)\rangle +\alpha^2\|\overline{\nabla F}(\M{X}^t)\|^2 +\alpha^2\Exp{\|\bar{\M{s}}^t\|^2\mid\mathcal F^t}.
\end{align*}
Using \eqref{eq:noise-history} and \eqref{eq:convex-average-inner} gives
\begin{equation}
\label{eq:convex-average-descent}
\Exp{\|\bar{\M{x}}^{t+1}-\M{x}^\star\|^2\mid\mathcal F^t} \le\|\bar{\M{x}}^t-\M{x}^\star\|^2-2\alpha[f(\bar{\M{x}}^t)-f^\star] +\frac{\alpha L}{n}\|\M{X}^t-\bar{\M{X}}^t\|_{\mathrm F}^2 -\alpha\left(\frac1L-\alpha\right)\|\overline{\nabla F}(\M{X}^t)\|^2 +\frac{\alpha^2\sigma^2}{n}.
\end{equation}
Define
\begin{equation}
\label{eq:distance-energy}
\mathcal L^t=\|\bar{\M{x}}^t-\M{x}^\star\|^2+\frac{4\alpha L}{g}U^t.
\end{equation}
Adding $4\alpha L/g$ times \eqref{eq:joint-consensus} to \eqref{eq:convex-average-descent} and using \eqref{UsefulInequality1}, we have
\begin{align*}
\Exp{\mathcal L^{t+1}\mid\mathcal F^t} &\le\mathcal L^t-2\alpha[f(\bar{\M{x}}^t)-f^\star] +\left(\frac{8\alpha L}{5}-\frac{4\alpha L}{g}\frac g2\right)U^t +\frac{\alpha^2\sigma^2}{n}+\frac{8\alpha^3L\sigma^2}{g}\\
&\quad+\left(-\alpha\left(\frac1L-\alpha\right)+\frac{100\alpha^5L^3}{p^2mg^2}\right) \|\overline{\nabla F}(\M{X}^t)\|^2.
\end{align*}
The coefficient of $U^t$ is $-2\alpha L/5$. The squared-gradient term is nonpositive since
\begin{align*}
\frac{100\alpha^5L^3}{p^2mg^2}\le\frac\alpha{4L} \le\alpha\left(\frac1L-\alpha\right).
\end{align*}
Consequently,
\begin{align*}
\Exp{\mathcal L^{t+1}\mid\mathcal F^t} \le\mathcal L^t-2\alpha[f(\bar{\M{x}}^t)-f^\star] +\frac{\alpha^2\sigma^2}{n}+\frac{8\alpha^3L\sigma^2}{g}.
\end{align*}
Taking full expectations and summing gives
\begin{align*}
\frac1T\sum_{t=0}^{T-1}\Exp{f(\bar{\M{x}}^t)-f^\star} \le\frac{R_0^2}{2\alpha T} +\frac{100\chi^2\alpha^2L\varsigma_0^2}{p^3(1-\lambda_2)^2T} +\frac{\alpha\sigma^2}{2n} +\frac{80\chi\alpha^2L\sigma^2}{p(1-\lambda_2)},
\end{align*}
which proves \eqref{CVXMainResult2-T}.
\end{IEEEproof}

\subsection{Convergence Analysis: Strongly Convex}\label{AP:strongly-convex}
\begin{IEEEproof}
In addition to \eqref{eq:convex-average-inner}, strong convexity and the smoothness upper bound at $\bar{\M{x}}^t$ imply
\begin{align*}
\langle\bar{\M{x}}^t-\M{x}^\star,\overline{\nabla F}(\M{X}^t)\rangle \ge f(\bar{\M{x}}^t)-f^\star -\frac L{2n}\|\M{X}^t-\bar{\M{X}}^t\|_{\mathrm F}^2 +\frac\mu2\|\bar{\M{x}}^t-\M{x}^\star\|^2.
\end{align*}
Here we used $\frac1n\sum_i\|\M{x}_i^t-\M{x}^\star\|^2\ge\|\bar{\M{x}}^t-\M{x}^\star\|^2$. Averaging this inequality and \eqref{eq:convex-average-inner}, and then expanding the squared distance update as above, gives
\begin{equation}\label{eq:strong-average-descent}
\begin{aligned}
\Exp{\|\bar{\M{x}}^{t+1}-\M{x}^\star\|^2\mid\mathcal F^t} &\le\left(1-\frac{\alpha\mu}{2}\right)\|\bar{\M{x}}^t-\M{x}^\star\|^2\\
&\quad-2\alpha[f(\bar{\M{x}}^t)-f^\star] +\frac{\alpha L}{n}\|\M{X}^t-\bar{\M{X}}^t\|_{\mathrm F}^2 -\alpha\left(\frac1{2L}-\alpha\right)\|\overline{\nabla F}(\M{X}^t)\|^2 +\frac{\alpha^2\sigma^2}{n}.
\end{aligned}
\end{equation}
Use the same energy \eqref{eq:distance-energy}. Since $\alpha\mu\le\alpha L\le g/16$, we have
\begin{align*}
\frac{8\alpha L}{5}+\frac{4\alpha L}{g}\left(1-\frac g2\right) \le\frac{4\alpha L}{g}\left(1-\frac{\alpha\mu}{2}\right).
\end{align*}
The forcing term is absorbed by \eqref{eq:forcing-absorption} and $\alpha L\le1/4$, because
\begin{align*}
\frac{100\alpha^5L^3}{p^2mg^2}\le\frac\alpha{4L} \le\alpha\left(\frac1{2L}-\alpha\right).
\end{align*}
Adding $4\alpha L/g$ times \eqref{eq:joint-consensus} to \eqref{eq:strong-average-descent} therefore yields
\begin{align*}
\Exp{\mathcal L^{t+1}\mid\mathcal F^t} \le\left(1-\frac{\alpha\mu}{2}\right)\mathcal L^t +\frac{\alpha^2\sigma^2}{n}+\frac{8\alpha^3L\sigma^2}{g}.
\end{align*}
Taking full expectations and iterating this inequality, we obtain
\begin{align*}
\Exp{\mathcal L^T} \le\left(1-\frac{\alpha\mu}{2}\right)^T\mathcal L^0 +\left(\frac{\alpha^2\sigma^2}{n}+\frac{8\alpha^3L\sigma^2}{g}\right) \sum_{t=0}^{T-1}\left(1-\frac{\alpha\mu}{2}\right)^t \le\left(1-\frac{\alpha\mu}{2}\right)^T\mathcal L^0 +\frac{2\alpha\sigma^2}{\mu n}+\frac{16\alpha^2L\sigma^2}{\mu g}.
\end{align*}
Since $\Exp{\|\bar{\M{x}}^T-\M{x}^\star\|^2}\le\Exp{\mathcal L^T}$, substituting \eqref{eq:energy-initialization} and \eqref{eq:proof-parameters} gives
\begin{align*}
\Exp{\|\bar{\M{x}}^T-\M{x}^\star\|^2} \le\left(1-\frac{\alpha\mu}{2}\right)^T \left(R_0^2+\frac{200\chi^2\alpha^3L\varsigma_0^2}{p^3(1-\lambda_2)^2}\right) +\frac{2\alpha\sigma^2}{\mu n} +\frac{320\chi\alpha^2L\sigma^2}{\mu p(1-\lambda_2)}.
\end{align*}
This proves \eqref{SCVXMainResult-NEW-T}.
\end{IEEEproof}

\subsection{Proof of Corollary \ref{MainResult-Convergence-rate-tighter-stepsize} and Corollary \ref{MainResult-Convergence-rate-C}}\label{SEC-PROOF-C}
Throughout this proof, write $\rho=1-\lambda_2$ and $h=p\rho/(320\chi L)$. The constants implicit in $\mathcal O(\cdot)$ below may depend on $L$, $\mu$ when $\mu>0$, and the displayed initialization quantities, but not on $p$, $n$, $\rho$, $\chi$, $\sigma$, or $T$.

\noindent\textbf{Non-convex and convex cases.}
We rewrite the non-convex bound established in the proof of Theorem \ref{MainResult-Convergence-rate-T} as
\begin{align*}
\frac1T\sum_{t=0}^{T-1}\Exp{\|\nabla f(\bar{\M{x}}^t)\|^2} \leq\underbrace{\frac{c_0}{\alpha T}+c_1\alpha+c_2\alpha^2}_{:=\Psi_T} +\frac{a_0\alpha^2}{T},
\end{align*}
where we may take
\begin{align*}
a_0=\frac{200\chi^2L^2\varsigma_0^2}{p^3\rho^2},\qquad c_0=1+2F_0,\qquad c_1=\frac{L\sigma^2}{n},\qquad c_2=\frac{160\chi L^2\sigma^2}{p\rho}.
\end{align*}
For the convex case, \eqref{CVXMainResult2-T} gives the same bound for $\frac1T\sum_{t=0}^{T-1}\Exp{f(\bar{\M{x}}^t)-f^\star}$, with
\begin{align*}
a_0=\frac{100\chi^2L\varsigma_0^2}{p^3\rho^2},\qquad c_0=1+\frac{R_0^2}{2},\qquad c_1=\frac{\sigma^2}{2n},\qquad c_2=\frac{80\chi L\sigma^2}{p\rho}.
\end{align*}
Suppose first that $\sigma>0$. Following the stepsize balancing argument in \cite[Lemma 17]{Stich2020}, choose
\begin{align*}
\alpha=\min\left\{\left(\frac{c_0}{c_1T}\right)^{1/2}, \left(\frac{c_0}{c_2T}\right)^{1/3},h\right\}.
\end{align*}
By this choice, we have
\begin{align*}
\frac{c_0}{\alpha T} \leq\frac{c_0}{hT}+\sqrt{\frac{c_0c_1}{T}}+\left(\frac{c_0^2c_2}{T^2}\right)^{1/3},\qquad c_1\alpha\leq\sqrt{\frac{c_0c_1}{T}},\qquad c_2\alpha^2\leq\left(\frac{c_0^2c_2}{T^2}\right)^{1/3}.
\end{align*}
Consequently,
\begin{align*}
\Psi_T+\frac{a_0\alpha^2}{T} \leq2\sqrt{\frac{c_0c_1}{T}} +2\left(\frac{c_0^2c_2}{T^2}\right)^{1/3} +\frac{c_0/h+a_0h^2}{T}.
\end{align*}
If $\sigma=0$, choose $\alpha=h$; the same inequality holds with $c_1=c_2=0$. Moreover, for the non-convex and convex cases, respectively,
\begin{align*}
\text{N-C: }\quad\frac{a_0h^2}{T} =\frac{200\varsigma_0^2}{320^2pT} \leq\frac{200\varsigma_0^2}{320^2}\frac{\chi}{p\rho T},\qquad \text{C: }\quad\frac{a_0h^2}{T} =\frac{100\varsigma_0^2}{320^2LpT} \leq\frac{100\varsigma_0^2}{320^2L}\frac{\chi}{p\rho T},
\end{align*}
where we used $0<\rho<2$ and $\chi\geq4$. Substituting the corresponding constants and $1/h=320\chi L/(p\rho)$ therefore gives, in both cases,
\begin{align*}
\Psi_T+\frac{a_0\alpha^2}{T} \leq\mathcal O\left(\frac{\sigma}{\sqrt{nT}} +\left(\frac{\chi\sigma^2}{p\rho T^2}\right)^{1/3} +\frac{\chi}{p\rho T}\right).
\end{align*}
Thus, \eqref{NCVXMainResult1-resp-new} and \eqref{CVXMainResult2-resp-new} hold for every $T\geq1$.

\noindent\textbf{Strongly convex case.}
Since $\alpha\leq h$, the initialization factor in \eqref{SCVXMainResult-NEW-T} satisfies
\begin{align*}
R_0^2+\frac{200\chi^2\alpha^3L\varsigma_0^2}{p^3\rho^2} \leq R_0^2+\frac{200\rho\varsigma_0^2}{320^3\chi L^2} \leq R_0^2+\frac{\varsigma_0^2}{L^2}=:a_0.
\end{align*}
Consequently, for some absolute constants $C_1,C_2>0$, Theorem \ref{MainResult-Convergence-rate-T} gives
\begin{align*}
\mathbb E\|\bar{\M{x}}^T-\M{x}^\star\|^2 \leq a_0 e^{-\alpha\mu T/2}+c_1\alpha+c_2\alpha^2,
\end{align*}
where $c_1=C_1\sigma^2/(\mu n)$ and $c_2=C_2\chi L\sigma^2/(\mu p\rho)$. If $\sigma>0$, define
\begin{align*}
\ell_T=\max\left\{1,\log\left(1+\frac{\mu a_0T}{2c_1}\right)\right\},\qquad \alpha=\min\left\{h,\frac{2\ell_T}{\mu T}\right\}.
\end{align*}
If $\alpha=h$, the initial term is $a_0e^{-\mu hT/2}$. Otherwise it is at most $a_0e^{-\ell_T}\leq2c_1/(\mu T)$. In both cases,
\begin{align*}
\mathbb E\|\bar{\M{x}}^T-\M{x}^\star\|^2 \leq a_0\exp\left(-\frac{\mu p\rho T}{640\chi L}\right) +\frac{4C_1\sigma^2\ell_T}{\mu^2 nT} +\frac{4C_2\chi L\sigma^2\ell_T^2}{\mu^3p\rho T^2}.
\end{align*}
This proves \eqref{SCVXMainResult-NEW-resp-new} with $c=\mu/(640L)$ in its exponential. Here the notation $\widetilde{\mathcal O}$ suppresses at most the explicitly displayed factors $\ell_T$ and $\ell_T^2$, with
\begin{align*}
\ell_T=\max\left\{1,\log\left(1+\frac{\mu^2a_0nT}{2C_1\sigma^2}\right)\right\}.
\end{align*}
If $\sigma=0$, choose $\alpha=h$. The bound becomes $a_0\exp(-\mu p\rho T/(640\chi L))$ directly; no logarithm involving $1/\sigma$ is used.

\noindent\textbf{Communication complexity.}
For the non-convex and convex cases, making each term in \eqref{NCVXMainResult1-resp-new} and \eqref{CVXMainResult2-resp-new} at most $\epsilon/3$ yields a sufficient deterministic horizon
\begin{align*}
T=\mathcal O\left(\frac{\sigma^2}{n\epsilon^2} +\sqrt{\frac{\chi}{p\rho}}\frac{\sigma}{\epsilon^{3/2}} +\frac{\chi}{p\rho\epsilon}\right).
\end{align*}
These expressions apply also when $\sigma=0$. The strongly convex complexity follows directly from the preceding fixed-stepsize bound. With the same $a_0,c_1,c_2$, set
\begin{align*}
\ell_\epsilon=\max\left\{1,\log\left(1+\frac{2a_0}{\epsilon}\right)\right\},\qquad \alpha=\min\left\{h,\frac{\epsilon}{4c_1},\sqrt{\frac{\epsilon}{4c_2}}\right\},\qquad T=\left\lceil\frac{2\ell_\epsilon}{\mu\alpha}\right\rceil.
\end{align*}
If $c_1=c_2=0$, their two stepsize bounds are interpreted as $+\infty$, so that $\alpha=h$. The initial term is at most $\epsilon/2$, and the two noise terms are each at most $\epsilon/4$. Hence the error is at most $\epsilon$, and
\begin{align*}
pT \leq p+\frac{2p\ell_\epsilon}{\mu} \left(\frac{1}{h}+\frac{4c_1}{\epsilon}+2\sqrt{\frac{c_2}{\epsilon}}\right).
\end{align*}
Substituting $h,c_1,c_2$ proves \eqref{SCVXMainResult-NEW}; the only suppressed logarithm in this communication bound is $\ell_\epsilon$. Finally, for the deterministic horizon $T$ chosen in each case, the expected number of communication rounds is exactly
$\mathbb E[\sum_{t=0}^{T-1}\theta_t]=pT$,
which proves \eqref{NCVXMainResult1}, \eqref{CVXMainResult2}, and \eqref{SCVXMainResult-NEW}. Integer rounding contributes at most one additional expected communication round and is absorbed into the stated bounds for sufficiently small $\epsilon>0$.

\end{document}